\documentclass[sigconf]{acmart}

\setcopyright{cc}
\setcctype{by}
\copyrightyear{2026}
\acmYear{2026}
\acmDOI{10.1145/3830398.3830722}
\acmConference[UIST '26]{The 39th Annual ACM Symposium on User Interface
  Software and Technology}{November 02--05, 2026}{Detroit, MI, USA}
\acmBooktitle{The 39th Annual ACM Symposium on User Interface Software and
  Technology (UIST '26), November 02--05, 2026, Detroit, MI, USA}
\acmISBN{979-8-4007-2856-3/2026/11}

\usepackage{subcaption}
\usepackage{graphicx}
\definecolor{DarkGreen}{HTML}{000000}
\definecolor{DarkBlue}{HTML}{000000}
\definecolor{PunchlineBlue}{HTML}{000000}
\newif\ifedittracking
\edittrackingfalse
\ifedittracking
  \definecolor{MajorEditColor}{HTML}{0000CC}   
  \definecolor{MinorEditColor}{HTML}{008000}   
\else
  \definecolor{MajorEditColor}{HTML}{000000}
  \definecolor{MinorEditColor}{HTML}{000000}
\fi
\newcommand\majoredit[1]{\textcolor{MajorEditColor}{#1}}
\newcommand\minoredit[1]{\textcolor{MinorEditColor}{#1}}
\newcommand\review[1]{#1}
\newcommand\minorreview[1]{#1}
\newcommand\punchline[1]{\textcolor{PunchlineBlue}{\textbf{#1}}}

\begin{document}


\title{Co-Annotator: Expert-Distilled ViT and VLM for Visual and Documentation Guidance in Age-Related Macular Degeneration}

\author{Ziheng~`Leo'~Li\textsuperscript{*}\quad Benjamin~Freeman\textsuperscript{*}\quad Akshay~Raman\textsuperscript{*}\quad Kavin~Aravindhan~Rajkumar\quad Xinxin~Fang\quad Rishabh~Srivastava\quad Steven~Feiner\quad Kaveri~A.~Thakoor}
\affiliation{%
  \institution{%
    Columbia University\\
    \{zihengleoli, feiner\}@cs.columbia.edu, \{kr3131, xinxin.fang, rs4489, k.thakoor\}@columbia.edu, \{baf2148, ar5126\}@cumc.columbia.edu\\[2pt]
    {\normalfont\footnotesize\textsuperscript{*}The first three authors contributed equally.}%
  }
  \country{}}


\renewcommand{\shortauthors}{Li, Raman, et al.}


\begin{abstract}
Clinical AI often optimizes predictive performance without engaging how clinicians decide where to look and what to write. We present \textit{Co-Annotator}, which distills expert gaze and dictation into two guidance components: a gaze-aligned Vision Transformer producing fixation-aligned areas of interest (AOIs), and an ontology-bounded vision--language model (VLM) that pre-fills editable biomarker summaries for retinal optical coherence tomography (OCT). We first collect expert gaze and dictations (US1) to train the models, significantly improving diagnostic accuracy and biomarker generation. We then deploy the system with ophthalmology residents: a controlled resident study (US2) confirmed each modality is safe and independently beneficial, with AOI guidance producing lasting perceptual efficiency gains through post-guidance carryover and VLM guidance more than doubling bio\-marker documentation breadth. In a combined deployment across two academic institutions (US3), providing both modalities simultaneously produced efficiency gains that substantially exceeded either modality alone: correct diagnoses per minute increased by 40\% and comment editing time fell by 67\%, without compromising diagnostic accuracy. \minoredit{Notably, neither modality improved efficiency \emph{during} guidance in US2, which makes the in-guidance efficiency gain under combined guidance in US3 the more striking result.} Expert-distilled multimodal guidance can remove two distinct clinical workflow bottlenecks at once (visual search overhead and documentation burden) without compromising the diagnostic accuracy clinicians already achieve.
\end{abstract}




\begin{teaserfigure}
  \includegraphics[width=\textwidth]{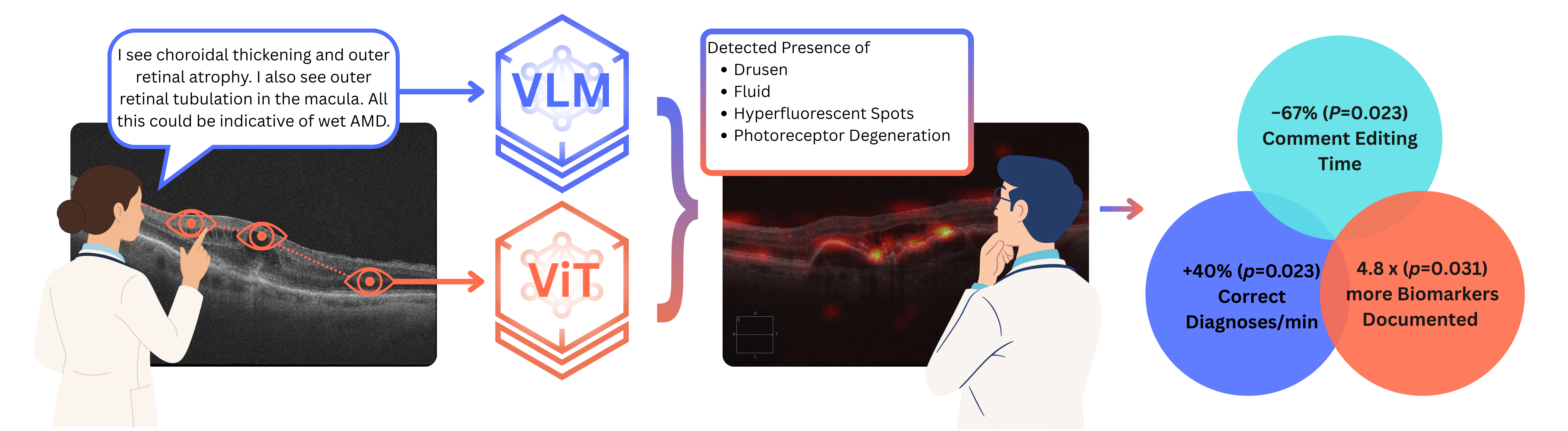}
\caption{\review{\textit{Co-Annotator} system architecture. Expert gaze and dictation train two backend models: a Vision Transformer (ViT) and a Vision--Language Model (VLM), powering two clinician-facing components: (1) fixation-aligned Areas of Interest (AOIs) that guide visual search, and (2) biomarker-bounded VLM guidance that scaffolds documentation.} Far right: combined US3 outcomes +40\% Correct Dx/min, $-$67\% comment editing time, and 4.8$\times$ more biomarkers documented (all $p<0.05$).}
\Description{Composite figure illustrating Co-Annotator: a clinician reviews an OCT scan while the system provides two guidance components powered by expert-distilled models. Left: a gaze-aligned ViT generates a semi-transparent AOI heatmap highlighting clinically relevant regions. Center: an ontology-bounded VLM pre-fills an editable biomarker summary in the comment field. Far right: three outcome circles from the combined US3 deployment showing +40\% Correct Diagnoses/min (p=0.023), minus 67\% comment editing time (p=0.023), and 4.8 times more biomarkers documented per AMD eye (p=0.031), all compared to unguided control.}
  \label{fig:teaser}
\end{teaserfigure}


\maketitle

\section{Introduction}\label{sec:intro}

We build and study an \minorreview{artificial intelligence} (AI) co-annotator \minorreview{for biomedical images} consisting of two physician-facing, vision--lan\-guage-model--backed user-interface components that highlight relevant imaging evidence to support trainees' diagnostic workflow. \minorreview{We specialize the system to diagnose wet age-related macular degeneration (wAMD) on optical coherence tomography (OCT) images.}
wAMD \minorreview{is an eye disease that} affects millions worldwide and is a leading cause of irreversible central vision loss in adults over 50 \cite{fleckenstein_age-related_2024}. Timely treatment can preserve vision when wAMD is identified before it causes permanent damage, yet early diagnosis is difficult: wAMD arises from growth of new abnormal blood vessels \minorreview{in the retina} causing fluid accumulation and bleeding, which must be resolved and distinguished from other AMD subtypes on OCT, the primary noninvasive imaging modality for diagnosing and monitoring \minorreview{wAMD} in routine practice \cite{metrangolo_oct_2021}. The relevant OCT biomarkers such as intraretinal fluid (IRF), subretinal fluid (SRF) or pigment epithelial detachment (PED) can be subtle and fine-grained, and they are challenging even for experienced readers to distinguish \cite{paez-escamilla_age-related_2021,saksens_macular_2014}.

Retina specialty clinics operate with high imaging volume and under significant time pressure. Each imaging study ordered by an ophthalmologist demands directed visual search, structured description, and auditable documentation. \minorreview{Additionally,} trainees are a critical user group because they show higher variance in accuracy and speed, are still forming mental models for biomarker patterns, and are calibrating trust in any computer-aided tools they use. Our system targets three workflow outcomes: (i) \textbf{faster, more focused reading} via area-of-interest (AOI) guidance , (ii) \textbf{lower documentation burden} via structured biomarker drafts residents can edit, and (iii) \textbf{greater transparency} by exposing where the model “looked” and constraining language to a clinical ontology.

Despite recent advances in vision and vision--language models (VLMs), faithful OCT biomarker grounding remains hard due to (i) domain shift across scanners, protocols, and patient populations; (ii) the need for \emph{spatially localized} distinctions (e.g., IRF vs.\ SRF) rather than global labels; and (iii) the risk of clinically unsafe hallucinations or vague free-text rationales \cite{li_artificial_2023,lim_vision_2024}. For safety-critical use, clinicians need transparency (to know where the model is attending visually), controllability (to be able to accept/edit findings), and outputs bounded by a shared ontology. These requirements motivate \emph{expert-aligned supervision} as well as interaction designs that present evidence to \emph{support} clinicians’ decisions rather than replace them.


We therefore built \review{\textit{Co-Annotator}, a system that unifies expert-distilled modeling with workflow-integrated design. The system consists of two backend models: a gaze-aligned vision transformer (ViT) and an ontology-bounded VLM. They drive two complementary clinician-facing interface components:}
{\setlength{\leftmargini}{1em}\setlength{\leftmarginii}{1em}
\begin{enumerate}
  \item \textbf{Fixation-aligned AOIs:} Attention overlays trained to visually guide readers toward regions with higher diagnostic relevance on OCT images, aligned to expert fixation density rather than post hoc saliency.
  \item \textbf{Biomarker-bounded VLM guidance:} A VLM fine-tuned on OCT images paired with expert dictations and curated biomarker labels to produce \emph{structured biomarker summaries} and answer targeted diagnostic questions.
\end{enumerate}}

To build and evaluate Co-Annotator, we conduct three user studies. \textbf{US1} collects synchronized expert gaze and dictations to train the ViT and VLM, establishing expert-distilled supervision targets. \textbf{US2} ($n{=}11$ ophthalmology residents) deploys each modality \emph{in isolation}, confirming guidance does not compromise residents' already-high diagnostic accuracy, and characterizing the individual benefit of each guidance channel. \minoredit{Efficiency gains in US2 emerged only \emph{after} guidance was removed, not during it.} \textbf{US3} ($n{=}8$ residents across Columbia University Irving Medical Center and Weill Cornell Medical Center) is our main contribution: it deploys both modalities \emph{simultaneously}, testing whether the combination produces gains neither modality achieves alone. Specifically, we make the following contributions:
{\setlength{\leftmargini}{1em}\setlength{\leftmarginii}{1em}
\begin{itemize}
  \item \majoredit{\textbf{Interaction design and design principles (primary contribution):} An interaction model for how clinicians and AI read medical scans together, distilled into three principles for clinical AI guidance that residents consistently demanded across all three studies: \emph{deferrable} (revealed after a first independent pass), \emph{sparse} (2--3 precise hotspots, not diffuse coverage), and \emph{evidence-anchored} (each biomarker token clickable to its AOI tile). The gaze-aligned ViT and ontology-bounded VLM are the enablers that let this guidance reflect how specialists actually look and describe findings, rather than the contribution in itself.}
  \item \textbf{Co-Annotator system:} An OCT co-annotation interface with two clinician-facing components designed for specificity, transparency, and editorial control: fixation-aligned AOI heatmaps (from a gaze-aligned ViT) and editable biomarker drafts (from an ontology-bounded VLM).
  \item \textbf{Expert distillation (US1):} The ViT and VLM faithfully align to expert visual attention and clinical language, improving diagnostic micro-AUC from 0.95 to 0.98 and achieving MedBERTScore 0.867 on biomarker text.
  \item \textbf{Modality evaluation (US2):} Each modality preserves residents' high baseline diagnostic accuracy while providing complementary benefits that establish the rationale for combining them: AOI guidance produces a post-guidance efficiency carryover and VLM guidance more than doubles biomarker breadth (5.8 vs.\ 2.3 per AMD eye, 83.1\% retention).
  \item \textbf{Combined deployment (US3):} Simultaneous AOI+VLM guidance significantly increased Correct Dx/min by 40\% ($p=0.023$), reduced comment editing time by 67\% ($p=0.023$), and yielded 5.36 biomarkers per AMD eye at 85.5\% retention. These outcomes exceed either modality alone (4.76 vs.\ 3.0 AOI-only and 2.0 VLM-only).
\end{itemize}}

\paragraph{Data and Model Availability.}
\label{Data and Model Availability}
The Expert Distillation Corpus from US1, the fine-tuned VLM, and the gaze-aligned ViT are publicly available.\footnote{Corpus: \url{https://huggingface.co/datasets/ramanakshay/amd-expert-corpus}; VLM: \url{https://huggingface.co/ramanakshay/vlm-oct-biomarkers}; ViT: \url{https://huggingface.co/ramanakshay/vit-oct-wamd}}

\section{Background}
\majoredit{\paragraph{Human-Centered Clinical AI}
Co-Annotator is fundamentally an interaction-design contribution: it concerns \emph{how} a clinician and an AI read a scan together, not only how accurate the underlying models are. This situates the work in a long line of HCI research on human--AI collaboration. Mixed-initiative interaction~\cite{horvitz1999principles} frames the AI as a partner that intervenes only when its contribution outweighs the interruption cost, motivating our \emph{deferrable} guidance. Trust calibration and human--AI complementarity, threads we return to below, further motivate our \emph{evidence-anchored} and \emph{sparse} design choices. Deployments of AI in real clinics, including diabetic-retinopathy screening in ophthalmology~\cite{beede2020human} and computational pathology~\cite{cai2019human}, show that workflow fit influences clinical value, beyond model accuracy alone. Our documentation component further builds on human--AI co-writing and clinical note-taking systems that keep the clinician in control of an editable AI draft~\cite{lee2022coauthor,yuan2022wordcraft,clark2018creative,gero2022sparks,buschek2021impact,murray2021medknowts,li2021automating,cai2021onboarding}. Building on this literature, we contribute three design principles for clinical AI guidance (\emph{deferrability}, \emph{sparseness}, and \emph{evidence-anchoring}) grounded in resident behavior across three user studies.}

\paragraph{Clinical context.}
Ophthalmic diagnosis for wAMD heavly relies on OCT and occurs at high clinical volumes. Subtle, fine-grained biomarkers drive treatment decisions yet are difficult to perceive reliably at speed: for example, distinguishing drusen (fatty deposits beneath the retina), pigment epithelial detachment (PED, a separation of the retinal pigment layer from its underlying membrane), and choroidal neovascularization (CNV, abnormal blood vessel growth) requires precise pattern recognition even for experienced clinicians~\cite{metrangolo_oct_2021,hanson_optical_2023,defauw2018clinically}. Decades of perception research show that expertise shapes visual search through distinct heuristics novices struggle to emulate~\cite{kundel2007holistic,van2017eyetrack}. While sharing expert gaze can aid detection~\cite{litchfield2010viewing}, direct replay is fragile: scanpaths are variable, unavailable for new cases, and hard to standardize~\cite{wu2019eye}. This motivates \emph{automated} AOI augmentation by training a model to predict fixation density for unseen images at scale.

\paragraph{Human–AI collaboration and trust calibration.}
There is broad consensus that clinical AI should support intermediate reasoning rather than replace judgment~\cite{castner2024expert,zhang2024rethinking,amershi2019guidelines,green2019principles}. Yet human–AI teams frequently fail to outperform the stronger agent alone due to miscalibrated trust and poor interaction design~\cite{vaccaro2024combinations}: participants may over-accept AI advice or be distracted without accuracy gains~\cite{bansal2021does,poursabzi2021manipulating,bussone2015role,yin2019understanding,lu2021human,laitan2019human,vasconcelos2023explanations}. Trust calibration, relying on AI more when it is right and discounting it when wrong, requires interaction structure, not just better explanations~\cite{lee2004trust,reverberi2022experimental,sakamoto2024facilitating,ghassemi2021falsehope,bucinca2021trust,bucinca2020proxy,zhang2020effect,wang2021explanations,schemmer2023appropriate,liao2020questioning,kocielnik2019accept,kaur2020interpreting,suresh2021beyond}.

\paragraph{Workflow fit and complementary roles.}
Workflow integration is the basis of accurate predictions. Previous studies have documented friction when AI outputs do not match clinicians' reading stages~\cite{kotter2021challenges,wenderott2024radiologists,tejani2024integrating,yang2019unremarkable,yang2016heartpump,wang2021brilliant,sivaraman2023ignore,jacobs2021designing,cai2019hello}. Conversely, designs for complementary roles, where AI handles one bottleneck while clinicians handle another, can exceed either alone in dermatology and pathology~\cite{tschandl2020human,kiani2020impact,inkpen2023advancing,rastogi2022deciding,xie2020chexplain,fogliato2022who,lee2021rehabilitation,panigutti2022understanding,yang2023harnessing}. Co-annotation interfaces that suggest \emph{where to look} and \emph{what to write}, communicate uncertainty, and preserve editorial control operationalize this principle~\cite{amershi2019guidelines,gomez2025human}.

\paragraph{Gaze-supervised vision models.}
Eye tracking captures diagnostic attention with low annotation burden and has been used as weak supervision to improve lesion detection and align model saliency with clinically meaningful regions~\cite{duan2025eye,wang2022follow,ibragimov2024use,wang2024gazegnn}. Multimodal methods use radiologists' gaze to align image features with report text~\cite{ma2024eye}, and ViT variants that incorporate expert gaze as an auxiliary input signal report concurrent gains in accuracy and interpretability~\cite{chen2026gaze}.

\paragraph{Ontology-bounded language models and our gap.}
LLMs in medicine require knowledge grounding to curb hallucinations: augmenting with Unified Medical Language System (UMLS)~\cite{umls} or structured report ontologies (e.g., RadGraph~\cite{jain2021radgraph}) constrains outputs to clinically valid concepts, while ontology-constrained decoding and retrieval-augmented prompting further reduce off-ontology drift~\cite{mehenni2024ontology,subramaniam2025ontology,yang2023integrating}. Despite this progress, most prior work optimizes gaze alignment~\cite{duan2025eye,ibragimov2024use} \emph{or} knowledge grounding~\cite{mehenni2024ontology,subramaniam2025ontology} in isolation, outside realistic reading and documentation loops. We unify both: fixation-aligned AOIs from expert gaze and an ontology-bounded VLM fine-tuned on expert dictations yield editable, evidence-forward guidance that fits the OCT workflow.

\majoredit{\section{Co-Annotator: Interaction Design}}
\label{sec:interaction-design}

\majoredit{\subsection{Embedded Costs of the Reading Loop}}
\label{sec:reading-loop}

\majoredit{Reading an OCT study is not a single classification act but a loop: the reader orients to the volume, searches across slices for structural abnormality, settles on a diagnosis, and then documents their findings for future reference. This loop incurs two distinct costs. The first is \emph{visual search overhead}: the biomarkers that matter are small, fine-grained, and distributed across five image slices per eye, so finding them is a scanning problem before it is a judgment problem. The second is \emph{documentation burden}: the finding must be written down in language consistent with a shared ontology, and that transcription competes with reading time, consuming a large share of per-eye time in our studies (\autoref{sec:us2}). Crucially, we treat these two costs as largely independent: reducing search time need not shorten transcription, and pre-filling text need not aid search. Co-Annotator therefore addresses them with two separate components rather than one general-purpose assistant, and whether that separation pays off is exactly what the combined deployment in \autoref{sec:us3} tests.}

\majoredit{\subsection{Two Guidance Components}}
\label{sec:components}

\begin{figure*}[t]
  \centering
    \includegraphics[width=\linewidth]{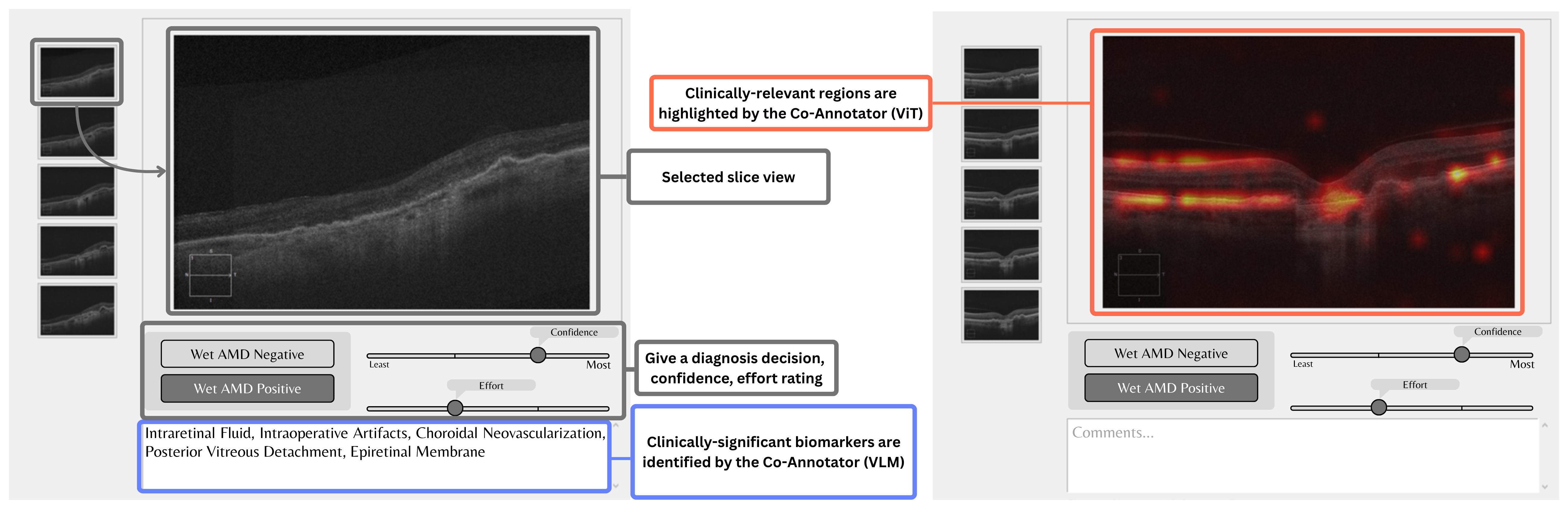}
    \caption{\review{Co-Annotator interface. Residents review five OCT slices per eye and give a binary diagnosis with confidence and effort ratings. \textbf{(a) VLM Guidance} (left): the biomarker comment field is pre-filled with VLM findings, which residents edit. \textbf{(b) ViT Guidance} (right): the interface with the gaze-aligned attention heatmap toggled on the OCT.}}
    \Description{Two side-by-side screenshots of the Co-Annotator interface. Left (VLM Guidance): an OCT retinal scan with a thumbnail strip, a biomarker text field pre-filled by the VLM, and confidence/effort rating controls. Right (ViT Guidance): the same interface with a warm-colored heatmap overlay highlighting clinically relevant regions.}
    \label{fig:system-uis}
\end{figure*}

Co-Annotator (\autoref{fig:system-uis}) provides two complementary components: a toggleable AOI heatmap overlay (from the gaze-aligned ViT) and an editable text box pre-filled with the VLM's biomarker draft for the majority diagnosis across the eye's five slices. The interface mimics routine clinical viewing conditions: five OCT slices per eye in a thumbnail strip, a radio-button diagnosis rating, and a free-text biomarker comment field.

\majoredit{The two guidance components share one screen and combine rather than compete: the AOI overlay renders directly on the OCT slice currently active in the thumbnail strip, beside the editable VLM comment draft. The overlay itself is rendered as a semi-transparent, colored heatmap rather than as outlines or contours around candidate regions; because the underlying B-scan (an individual slice of the OCT volume) is grayscale, a colored, semi-transparent layer remains legible against it without obscuring the scan beneath. The overlay's visual parameters (color map and transparency level) were inherited from our prior glaucoma diagnosis interface~\cite{li2025interactively} and were set empirically through informal design discussions with clinicians rather than tuned for this study. The overlay is toggleable by right-clicking the image, allowing them to move fluidly between an unguided first look and a guided check. The client itself is a lightweight, Python-based interface built for this logging and toggling behavior.}

\paragraph{Design Rationale as Built}
\label{sec:design-rationale}
\majoredit{Two principles bear directly on the design of the interface. Guidance is \emph{dismissible} rather than ambient: because the overlay is toggled by the reader, an unguided first pass is always available, which was apparently preferred by residents (\autoref{sec:us2}). Guidance is also \emph{bounded and editable} rather than authoritative: the draft comment can only use terms from the expert-derived ontology, and is subjected to the clinician's review. We are deliberately narrow about two other properties.
We expect to implement a \emph{sparse} overlay that surfaces only a few hotspots and to link individual biomarker tokens to the image regions that produced them. 
Residents asked for both, repeatedly, and we report those requests as findings in \autoref{sec:us3-qual} and~\autoref{sec:discussion} rather than as validated design decisions.}

\majoredit{\section{Expert-Distilled Models}}
\label{sec:models}
\majoredit{The two models in this section enable the interaction design above: we distill expert gaze and dictation so that the AOI and biomarker guidance reflect how specialists actually look and describe findings.} \review{Development proceeds in three stages: data collection (\autoref{sec:expert-distillation-data}), model training (\autoref{sec: vit} and \autoref{sec:vlm}), and interface integration (\autoref{sec:interaction-design}). Because rich OCT diagnostic-reasoning datasets are scarce, we build our own corpus in US1, and because the expert pool is small we train with a two-stage hierarchical curriculum for efficient transfer.} \majoredit{Readers who are interested in using the model and the dataset can refer to \autoref{Data and Model Availability}. Model and training details are in \autoref{appendix:vlm-training} and \autoref{appendix:vit-training}; we include here only what is needed to understand the guidance residents see.}

\subsection{Expert Knowledge Distillation}
\label{sec:expert-distillation-data}

Experts’ eye movements and spoken dictations were transformed into the \emph{supervision targets} that underpin both models: fixation-density maps for the gaze-aligned ViT and ontology-bounded bio\-marker labels for the VLM.

\subsubsection{In-house Dataset}
\label{sec: in-house dataset}
We collected \textbf{1{,}155} high-definition five-line raster OCT scans of the macula from \textbf{231} eyes in \textbf{203} patients (five scans per eye; \textbf{104} normal, \textbf{127} wAMD). \review{Scans were captured on the Zeiss Cirrus OCT imaging platform, which is likely the most commonly used for retinal imaging \cite{Akman2018}.} All data were collected in accordance with the principles laid out in the Declaration of Helsinki under a protocol approved by our Institutional Review Board, and de-identified according to national law. We will refer to this dataset as the \textit{in-house dataset}. \review{We collected visual attention from eight experts on a subset of 138 images, and dictation on a subset of 113 (see \autoref{sec:us1})}. 

\subsubsection{Visual Attention}
Eye movements recorded during expert OCT reads were converted into a per-image fixation-density target $A^*$ on the $32{\times}32$ ViT patch grid (fixation detection, density estimation, hardware, and calibration in \autoref{appendix:vit-training}). \review{$A^*$ represents cumulative expert attention over the full diagnostic review, linking gaze to the holistic diagnosis rather than individual biomarker tokens.} These maps directly compare with attention rollout on the same patch lattice (\autoref{sec: vit}).

\subsubsection{Expert Dictation and Clinical Biomarkers}
Audio was recorded continuously while experts examined images in the \emph{Single Image Input} setting and was time-synced to the active image. Recordings were transcribed to text and lightly cleaned by experts to remove noise while preserving clinical content. 

Free-text dictations were converted to multi-label biomarker targets using an ontology derived from OCT5k~\cite{arikan2025oct5k}: a prompted extraction model identified presence cues from transcripts and output only terms from an exclusive ontology list, with outputs reviewed by a retina specialist (full extraction procedure in \autoref{appendix:dataset}). The in-house corpus was enhanced with OCT5k samples to form 573 image--biomarker pairs (distribution in \autoref{appendix:distribution}).



Biomarkers with fewer than 10 occurrences were excluded, yielding 12 predefined terms (listed in \autoref{appendix:dataset}) used as the ontology for VLM training and evaluation.

This phase of US1 ultimately produced the Expert Distillation Corpus: (1) patch-aligned attention targets $A^*$ per image for gaze-aligned ViT supervision, and (2) ontology-bounded biomarker multi-labels for VLM fine-tuning and evaluation.

\subsection{AOI Localization with Gaze-Aligned ViT}
\label{sec: vit}

\review{To ensure clinically grounded visual saliency, we trained a ViT to align its attention to expert fixations. At each step, the ViT produces a classification prediction $\hat{y}$ and, via attention rollout across all transformer layers, an attention map $\hat{A}^{(L)}(x)$;} \minoredit{training then minimizes a classification term plus a gaze-alignment term weighted by $\alpha$:}
\review{\begin{equation*}
\mathcal{L} \;=\; \mathrm{BCE}(\hat{y},\, y) \;+\; \alpha \cdot \mathrm{CrossEntropy}\!\big(\hat{A}^{(L)}(x),\, A^*\big).
\end{equation*}}
\minoredit{Raising $\alpha$ pulls the network's attention toward the regions experts fixated; the classification term preserves diagnostic performance.}

\review{Crucially, this objective lets the ViT learn a generalized mapping between visual features (e.g., fluid pockets and hyperreflective spots) and expert attention: at inference, it predicts fixation-aligned AOIs for \emph{unseen} images, generating guidance for new patients without requiring expert eye-tracking.} Loss definitions, the $\alpha$ sweep, architecture, the attention rollout algorithm, training configuration, and ablations are in \autoref{appendix:vit-training}.


\subsection{Two-Stage VLM for Biomarker Suggestion}
\label{sec:vlm}
Having aligned the ViT’s visual attention to expert gaze, we now describe the complementary language component. We fine-tune MedGemma \cite{sellergren2025medgemma}, a large-scale VLM pretrained on diverse medical datasets, to specialize it at interpreting ophthalmic OCT scans. Our framework produces a single, unified model trained to perform three tasks: diagnosis, biomarker discrimination, and biomarker identification (\autoref{tab:vlm_tasks}). To ensure clinical utility, the model interaction is governed by a structured prompting methodology to minimize ambiguity and a strict overarching system prompt (\autoref{tab:vlm_tasks}).
We designed a two-staged curriculum to help the model build a robust and domain-specific visual foundation for the first two tasks, before attempting the more challenging third task of generating biomarkers.

\begin{table*}[t]
\centering
\caption{VLM tasks, objectives, outputs, and prompt placeholders.}
\label{tab:vlm_tasks}
\begin{tabular}{p{0.14\textwidth} p{0.3\textwidth} p{0.18\textwidth} p{0.28\textwidth}}
\toprule
\textbf{Task} & \textbf{Objective} & \textbf{Output Format} & \textbf{Prompt} \\
\midrule
Diagnosis & Binary classification of the OCT scan’s primary pathology. & \texttt{Normal} or \texttt{wAMD} & What is the diagnosis for this OCT scan, Normal or wAMD? \\
Biomarker\\Discrimination & Verify presence of a queried biomarker (VQA-style). & \texttt{Yes} or \texttt{No} & Is \textless BIOMARKER\textgreater{} present in this OCT scan? \\
Biomarker\\Generation & List all present biomarkers from the ontology, or none. & Comma-separated ontology terms or \texttt{None} & What biomarkers are visible in this OCT scan? \\
\addlinespace[2pt]
\multicolumn{4}{l}{\footnotesize All tasks share a structured system prompt constraining output format and domain (see \autoref{appendix:vlm-prompt}).}\\
\bottomrule
\end{tabular}
\Description{A detailed table describing the VLM tasks parameters, including task, objective, desired output format, and the prompt employed.}

\end{table*}


\paragraph{Two-stage curriculum.}
Stage one trains exclusively on 1) diagnosis: binary classification, OCT-C8 + in-house data, and 2) biomarker discrimination: visual question-answering (VQA)-style yes/no que\-ries with the Expert Distillation Corpus. The goal is to build OCT-domain grounding before tackling the harder generative task. Stage two adds biomarker generation \minoredit{while continuing to train on the first two tasks, so the model keeps its diagnostic and discrimination grounding while learning to generate ontology-bounded biomarker labels (dataset compositions in \autoref{appendix:dataset}; sampling strategy and hyperparameters in \autoref{appendix:vlm-training})}.

\section{User Study 1 (US1): Distilling Expert Knowledge via Gaze and Dictation}
\label{sec:us1}\label{sec:past_aoi_exp}
US1 evaluates whether expert behavioral signals (visual fixations and concurrent dictations) can be faithfully captured and distilled into (i) a gaze-aligned ViT and (ii) an ontology-bounded VLM that generalizes to unseen clinical cases. Eight retina specialists provided gaze data in Setup A (five-image bundles, no dictation) and dictations in Setup B (single image, concurrent narration).


\label{desc:attention-faithfulness-hypothesis}\label{desc:ontology-grounding-hypothesis}

\subsection{Results and Discussion}
The gaze and dictation corpus underlying US1 (and its expansion with OCT5k~\cite{arikan2025oct5k} to the final sample set) is described in \autoref{sec:expert-distillation-data}.

\subsubsection{Gaze Alignment Improves Diagnostic Accuracy}
To evaluate 
US1, we computed the diagnostic accuracy of the ViT with its attention biased toward clinically meaningful regions, using the auxiliary \textit{expert-alignment loss} (\autoref{sec: vit}). We did so across a range of alignment weights, $\alpha$, to vary the extent to which expert fixation factors in the ViT training. Because each eye's scan has five corresponding layers, it is important to test 
US1 in two input regimes: \textbf{patient-level (multi-image)}, where all five OCT images per eye are concatenated and jointly processed to mirror what a clinician sees; and \textbf{single-image}, where each image with expert visual attention data is treated as an independent sample.

\begin{figure*}[t]
  \centering
  \includegraphics[width=\textwidth]{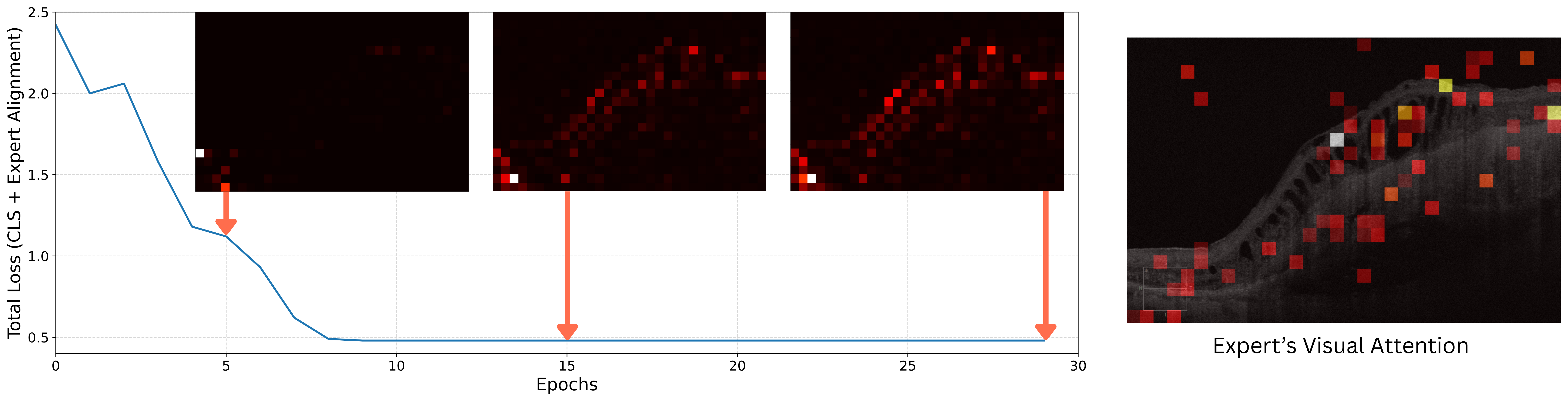}
  \caption{Gaze-aligned ViT attention converges to expert fixations during training. Insets: last-layer attention rollout at epochs 5, 15, and final vs.\ target $A^*$ (expert fixation density). Attention begins diffuse and sharpens onto clinically relevant structures as loss decreases.}
  \label{fig:aoi vit results}
  \Description{A line graph with inset images showing total loss versus number of epochs. Loss decreases with the number of epochs. Insets display the ViT’s last-layer attention rollout (normalized on the $32\times32$ patch grid) across epochs (5, 15, and final) for the same OCT image; the right panel is the expert visual attention $A^*$ derived from their gaze data. Early attention is diffuse and off-target; as training progresses, the model’s attention concentrates on the same structures emphasized by experts, while the loss improves.}
\end{figure*}

For the \textit{patient-level input}, attention alignment consistently improved diagnostic accuracy. Adding gaze alignment raised \textbf{micro-AUC from 0.95 to 0.98} with five-fold cross-validation. The amount of data is limited under this setting, as not all images in each set of five had expert visual attention data. Thus, we focus on the results in the \textbf{single–image} setting. The best trade-off occurred at \(\alpha=0.05\), yielding \textbf{88.37\%} validation accuracy and \textbf{F1 = 0.8628}, outperforming the baseline ViT without alignment (\autoref{tab:vit results}). The improved classification supports
US1. Qualitatively, as in \autoref{fig:aoi vit results}, model saliency resembled expert fixation distributions, which is consistent with the intuition that experts' visual attention can cue the model to attend to clinically relevant regions.

\subsubsection{Dictation-Derived Biomarker Generation}

We evaluated the efficacy of the two-stage VLM training across the three tasks described in \autoref{sec:vlm}. The model achieved high accuracy on discriminative tasks. Diagnostic accuracy was \textbf{0.920} on the OCT-C8 held-out test set and \textbf{0.910} on a larger 
US1 test set (\autoref{tab:vlm-results}). Biomarker discrimination had an accuracy of \textbf{0.800}. (\autoref{tab:vlm-results}).  For generative biomarker identification, the model attained strong semantic fidelity to dictation-derived references \minorreview{(\textbf{BERTScore$_{\text{F1}}$ = 0.880}, \textbf{MedBERTScore$_{\text{F1}}$ = 0.867})}. An ablation study revealed that integrating OCT-5K with our Expert Distilled Corpus was critical: training only on the smaller transcript dataset led to marked overfitting. 

Taken together, these results support that the dictation-derived, ontology-bounded biomarker targets served as effective supervision and assessment signals for biomarker text generation, yielding high semantic similarity while maintaining strong discriminative performance. Together, these results motivate the deployment of Co-Annotator with ophthalmology residents to evaluate whether expert-distilled AOI and VLM guidance improve clinical workflow outcomes in US2 and US3.

\begin{table*}[tp]
\centering
\caption{Performance of the Fine-Tuned MedGemma Model on all three tasks after the two-stage training. To gauge generalization, diagnosis performance is evaluated on OCT-C8 \cite{subramanian_classification_2022} and US1 data. Base refers to the pretrained MedGemma without fine-tuning.}
\label{tab:vlm-results}
\begin{tabular}{@{}lllrrr@{}}
\toprule
\textbf{Task Category} & \textbf{Task} & \textbf{Metric} & \textbf{\shortstack{fine-tuned on \\OCT5K + US1}} &
\textbf{\shortstack{fine-tuned on\\US1}} & \textbf{Base} \\
\midrule
Binary Choice & Diagnosis (US 1) & Accuracy & 0.910 & 0.800 & 0.290 \\
 & Diagnosis (OCT-C8 \cite{subramanian_classification_2022}) & Accuracy & 0.920 & 0.900 & 0.400 \\
 & Biomarker Discrimination & Accuracy & 0.800 & 0.840 & 0.535 \\
\midrule
Generative & Biomarker Generation & BERT Score (F1) & 0.880 & 0.818 & 0.790 \\
 & & MedBERT Score (F1) & 0.867 & 0.774 & 0.698 \\
\bottomrule
\end{tabular}
\Description{A table reporting the performance of our fine-tuned MedGemma model on all three tasks. The model generated the best diagnostic performance and biomarker outputs when fine-tuned on the combination of OCT5K and our in-house dataset (accuracy 0.92 on OCT-C8 and 0.91 on US1 corpus, MedBERTScore 0.867, BERTSCore 0.880), better than when fine-tuned only on US1 and better than baseline. }

\end{table*}

\section{User Study 2 (US2): Controlled Evaluation of Individual Guidance Modalities}
\label{sec:us2}

Building on US1, we evaluated each guidance modality, AOI heat\-maps (\emph{where to look}) and VLM text drafts (\emph{what to write}), in isolation before combining them. Separate experiments (i) isolate causal mechanisms and avoid interaction confounds, (ii) confirm guidance does not compromise accuracy, and (iii) keep sessions clinically feasible. The key hypotheses were:

{\setlength{\leftmargini}{1em}\setlength{\leftmarginii}{1em}
\begin{description}
  \item[H2.1\label{accuracy-hypothesis}] \textbf{(Clinical safety)} Each modality preserves diagnostic accuracy. Either guidance does not harm clinical performance.
  \item[H2.2\label{us2-perceptual-hypothesis}] \textbf{(Perceptual guidance)} AOI heatmaps support more efficient visual search and encode lasting viewing strategies that carry over after guidance is removed.
  \item[H2.3\label{us2-documentation-hypothesis}] \textbf{(Documentation assistance)} VLM drafts broaden biomarker documentation breadth while residents retain editorial agency, aligning vocabulary with the clinical ontology and filtering off-ontology suggestions.
\end{description}}

\subsection{Participants and Study Design}\label{us2-setup}
Residents were recruited from a pool of 16 ophthalmology trainees across academic medical training programs. Power analyses targeted the detection of a 0.10 absolute accuracy difference at the eye level and medium effects on timing and behavioral measures (Cohen’s $d=0.5$), translating to image-set sample goals of $n{=}74$ for accuracy-focused tests and $n{=}128$ for other outcomes, achievable with 5–9 residents reading $\sim$15 eyes per block. Ten residents completed the AOI study and four completed the VLM study (four overlapped), totaling 11 unique participants, representing 69\% of the available pool. The VLM sub-study ($n{=}4$ participants, 120 image-set observations) was sized for the biomarker outcome ($n{=}128$ image-set target), not for timing outcomes; participant-level timing estimates for the VLM condition should be interpreted as exploratory. 
\textit{Interface.}
\majoredit{Residents read OCT scans using the Co-Annotator interface described in \autoref{sec:interaction-design} (\autoref{fig:system-uis}).} The interface logs all interactions with millisecond timestamps\majoredit{, and} participants could not return to previously rated cases to ensure independent reads.

\textit{Apparatus.}
All sessions were conducted on a standard research workstation with a $1920{\times}1080$ monitor at 60\,Hz; residents interacted via keyboard and mouse. In the US2 AOI condition, eye movements were recorded with a \textit{Tobii Pro Fusion} tracker sampling at 250\,Hz, mounted on the monitor bezel. A standard 9-point calibration was performed at the start of each session (maximum accepted error $<0.5^\circ$). \minoredit{Eye tracking was recorded in the US2 AOI condition and in US3; it was not used in the US2 VLM condition.}

\begin{figure}[h]
  \centering
  \includegraphics[width=\linewidth]{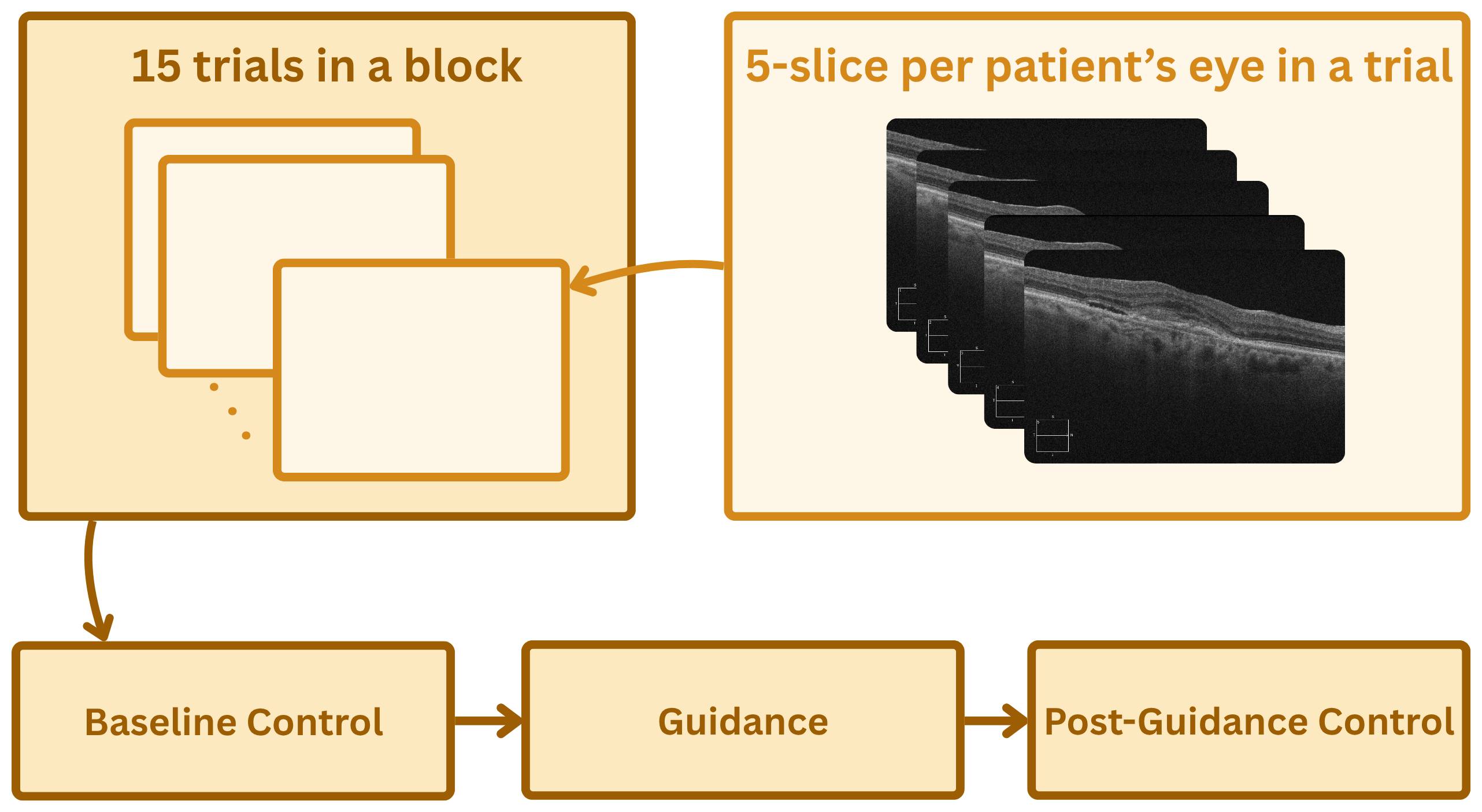}
  \caption{Study flow: each trial presents a patient's 5-slice OCT volume; 15 trials form a block. The AOI study followed Baseline Control $\rightarrow$ Guidance $\rightarrow$ Post-Guidance Control.}
  \Description{A flowchart showing the study block structure. Each trial presents 5 OCT slices for one patient eye. 15 trials form one block. The AOI study block sequence is Baseline Control, then Guidance, then Post-Guidance Control.}
  \label{fig:userstudylayout}
\end{figure}

\textit{Study protocol.}
Each experiment included unguided control blocks (standard-of-care baseline) alongside guidance blocks. The AOI condition followed a three-block sequence: pre-guidance control $\to$ AOI guidance $\to$ post-guidance control (\autoref{fig:userstudylayout}), enabling measurement of carryover effects. The VLM condition used two blocks (pre-control $\to$ VLM guidance) to bound session length to under 40 minutes. All sessions began with a short practice run.

Primary outcomes were diagnostic accuracy, Correct Dx/min, time per eye, and comment edit time; biomarker retention and vocabulary overlap (Jaccard index) were additionally tracked in the VLM condition. Statistical methods, outcome variable definitions, and ordering effect analyses are detailed in \autoref{appendix:stats}.


\begin{figure}[h]
  \centering
  \includegraphics[width=\linewidth]{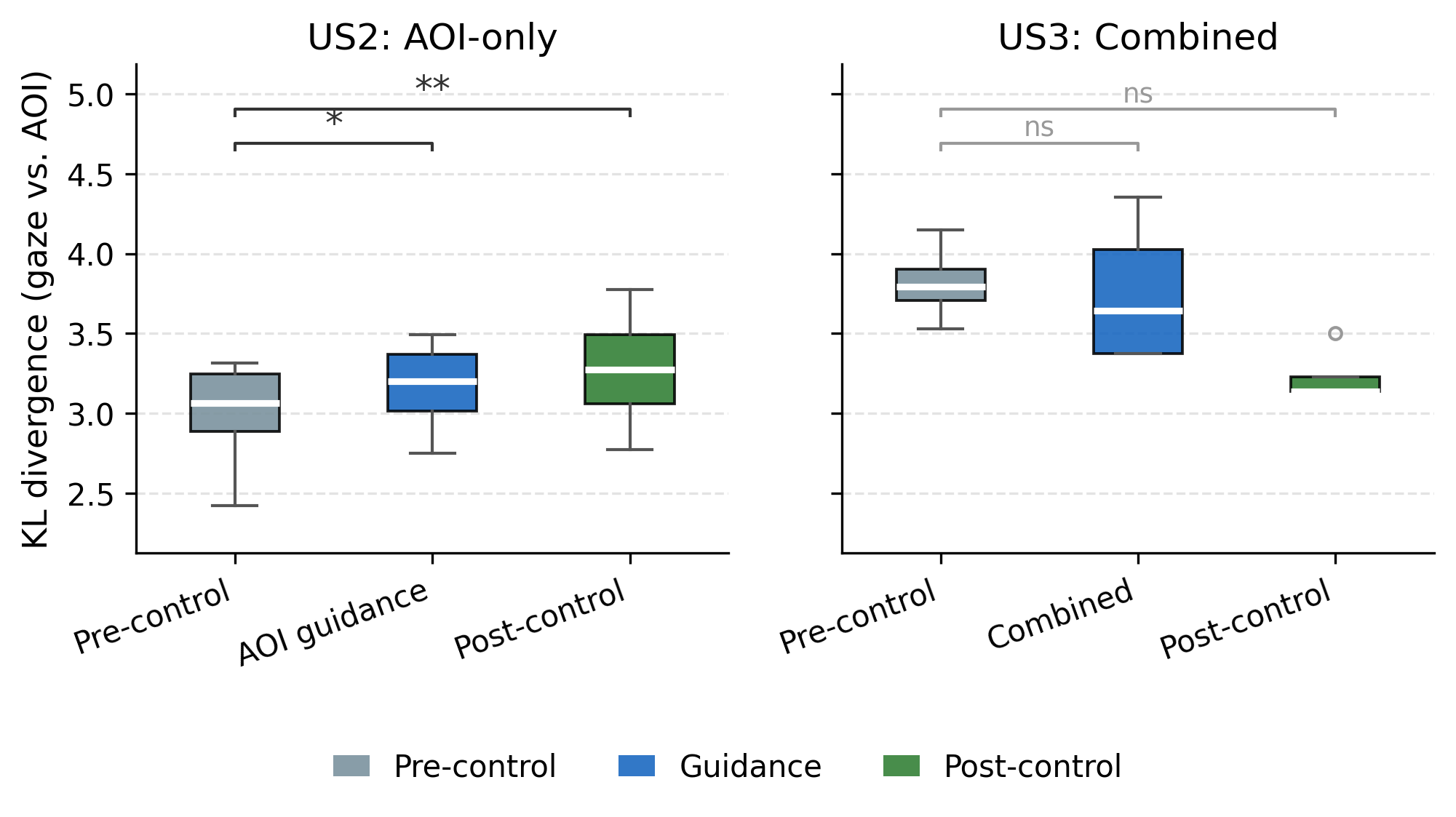}
  \caption{Kullback--Leibler (KL) divergence (gaze vs.\ model AOI) by block for US2 AOI-only (left) and US3 combined guidance (right). AOI-only guidance significantly increased divergence during and after guidance (* $p{<}0.05$, ** $p{<}0.01$ vs.\ pre-control); combined guidance showed no significant change (all $p{>}0.6$).}
  \Description{Two-panel boxplot with connected per-participant lines. Left (US2 AOI-only): divergence increases significantly during guidance (p=0.039) and post-guidance (p=0.008). Right (US3 combined): divergence is flat across all blocks with no significant differences.}
  \label{fig:divergencefigure}
\end{figure}


\subsection{Results and Discussion}

\noindent\textbf{AOI guidance.} Diagnostic accuracy was preserved across all blocks (H2.1; \autoref{tab:all_results}). In-guidance efficiency gains (H2.2) were not observed: Correct Dx/min and time per eye were statistically unchanged during the AOI block relative to pre-control. This null result is interpretable; residents adopted a read-first strategy, toggling the heatmap only after forming an initial impression, so any benefit would surface after the unaided pass rather than during it. The post-guidance block showed significant carryover in time per eye, which fell 28\% versus pre-control ($p{<}0.001$); Correct Dx/min rose 52\% but did not reach significance ($p{=}0.065$ raw, $0.274$ adjusted), supporting H2.2, though task familiarity cannot be fully ruled out given the fixed block order. \majoredit{Because diagnostic accuracy was preserved across all three blocks (0.93/0.89/0.90), Correct Dx/min here tracks reading speed: the post-guidance rise (2.9$\rightarrow$4.4, $+52\%$) is the counterpart of the time-per-eye carryover (residents read faster unaided after internalizing where to look), while the in-guidance null follows from the read-first-then-toggle pattern, where consulting the overlay adds a step that offsets any speed gain during the guided block.} Residents spent 17--22\% of per-eye time editing comments, a documentation bottleneck AOI guidance alone does not address.

\noindent\textbf{VLM diagnostic guidance.} Diagnostic accuracy was maintained (H2.1; 0.917 vs.\ 0.833 in control, nonsignificant); no statistically significant change was observed in timing or effort outcomes (\autoref{tab:all_results}).


\noindent\textbf{VLM biomarker guidance.} VLM guidance significantly broadened documentation: residents recorded a mean of 5.8 biomarkers per AMD eye versus 2.3 in control ($p{=}8.8\!\times\!10^{-9}$), retaining 83.1\% of VLM suggestions while deleting 16.8\% and independently contributing 27 new biomarkers (\autoref{fig:us2-biomarkers}). Vocabulary overlap (Jaccard) increased from 0.45 to 0.78. Together these support \textbf{H2.3}. Hallucination rate was low: 3 out-of-ontology biomarkers across 11 appearances; the high-risk case (``hemorrhage’’) was deleted 71\% of the time. Full $p$-values and mixed-effects estimates are in \autoref{appendix:stats}.


Critically, neither modality improved efficiency \emph{during} its guidance block (H2.2): the benefits were indirect: an AOI post-guidance timing carryover and a VLM documentation-only gain, each already detailed above. This is what makes the US3 in-guidance gain of $+40\%$ Correct Dx/min ($p=0.023$) notable: it is a gain neither modality produced alone during active guidance, and it motivates combining the two, each targeting a distinct bottleneck.

\section{User Study 3 (US3): Combined VLM+AOI Guidance}
\label{sec:us3}

US2 established that AOI guidance benefits perceptual efficiency and VLM guidance benefits documentation completeness. US3 tests whether their combination produces gains exceeding either modality alone:

{\setlength{\leftmargini}{1em}\setlength{\leftmarginii}{1em}
\begin{description}
  \item[H3.1\label{us3-accuracy-hypothesis}] \textbf{(Accuracy preservation)} Combined guidance maintains diagnostic accuracy compared to unguided reading.
  \item[H3.2\label{us3-efficiency-hypothesis}] \textbf{(In-guidance efficiency)} Combining AOI and VLM guidance increases Correct Dx/min and reduces time per eye and comment edit time relative to the pre-guidance control.
  \item[H3.3\label{us3-carryover-hypothesis}] \textbf{(Post-guidance carryover)} Efficiency gains persist in the post-guidance block.
  \item[H3.4\label{us3-biomarker-hypothesis}] \textbf{(Biomarker breadth)} Combined guidance achieves bio\-mark\-er breadth $\geq$ that of US2 VLM-only condition.
  \item[H3.5\label{us3-retention-hypothesis}] \textbf{(VLM retention)} Residents retain $\geq$70\% of VLM-suggested biomarkers in the combined interface.
\end{description}}

\subsection{Participants and Study Design}
\label{sec:us3-setup}

Eleven ophthalmology residents were recruited across two academic medical institutions; eight completed the study (median residency training: 22 months). Residents who had participated in US2 were eligible with a six-month washout to minimize direct task recall. Of the eight completers, three had previously participated in US2 (returning from the first institution) and five were new recruits from the second institution. This partial overlap is a potential confound for cross-study comparisons, though the washout period and within-subject block design mitigate direct carryover of specific image responses. Power analysis followed US2 targets ($n{=}74$ image sets for accuracy, $n{=}128$ for timing, achievable with eight residents reading 15 eyes per block). Recruiting across two institutions improved ecological validity and reduced site-specific confounds while preserving the within-subject design.

The study followed the same three-block design as the US2 AOI experiment (pre-guidance control $\to$ combined guidance $\to$ post-guidance control), with 15 eyes per block, using the same interface (\autoref{sec:interaction-design}). In the guidance block, residents received both modalities simultaneously (\autoref{sec:interaction-design}). A post-study survey ($n{=}8$, 1–7 Likert) assessed perceived accuracy of guidance, learning carryover, reliance, and preference for continued access.

The primary outcome was Correct Dx/min; secondary outcomes matched US2 (\autoref{tab:all_results}). Biomarker retention and vocabulary overlap were computed identically to US2. Ordering effects were confirmed nonsignificant for accuracy via the same practice-slope decomposition (\autoref{appendix:stats}).

\subsection{Results and Discussion}

\subsubsection{Diagnostic Accuracy and Efficiency}

\autoref{tab:all_results} summarizes results across the three blocks. Residents' diagnostic accuracy was not compromised under combined guidance, and if anything higher ($83.3\%$ pre-control vs.\ $93.3\%$ under guidance; $W{=}2$, $n{=}8$, $p=0.094$; \autoref{tab:all_results}); false positive rate remained 0\% throughout. The guidance made residents faster and not at the expense of the accuracy they already had (H3.1).

\begin{figure*}[t]
  \centering
  \begin{subfigure}{0.62\textwidth}
    \includegraphics[width=\linewidth]{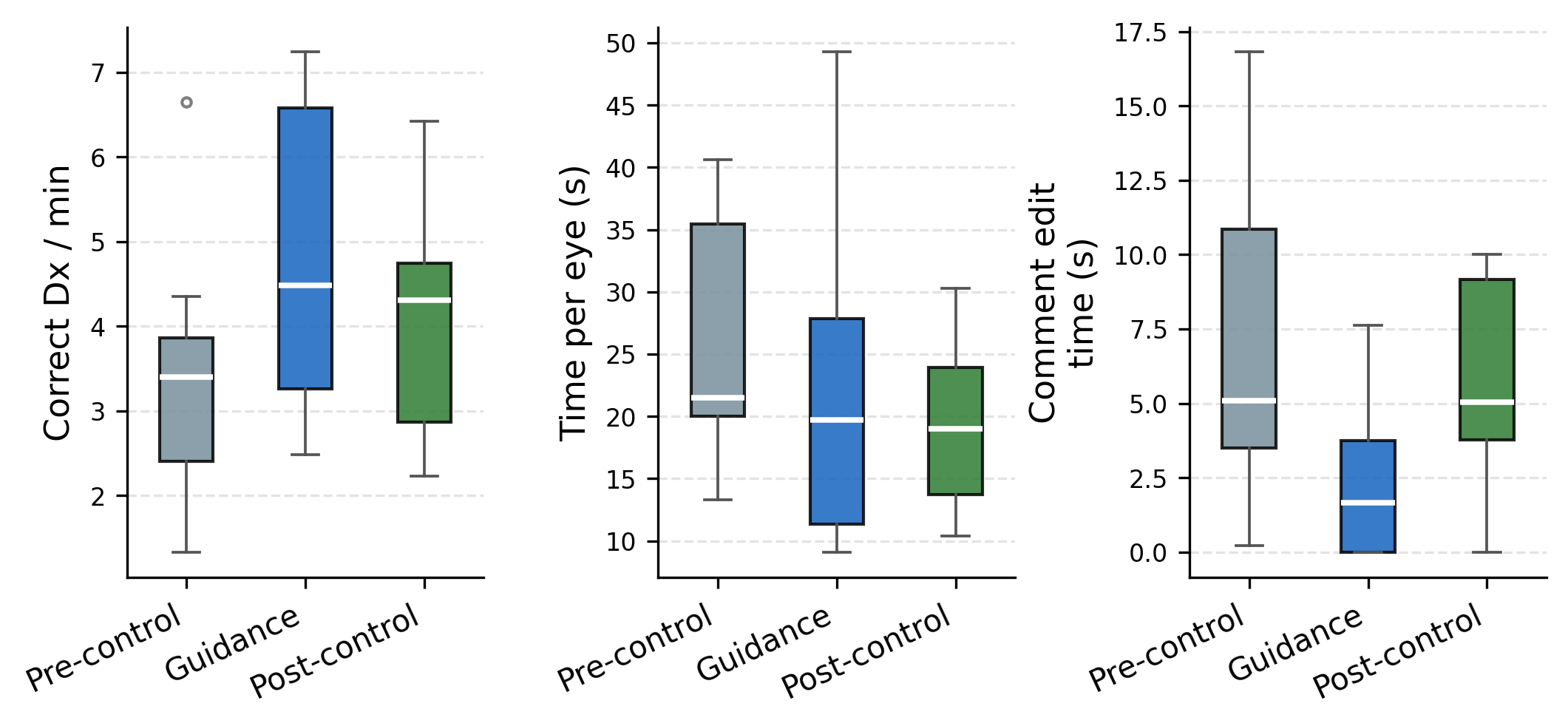}
    \caption{Per-block distributions with connected participant lines.}
  \end{subfigure}
  \hfill
  \begin{subfigure}{0.36\textwidth}
    \includegraphics[width=\linewidth]{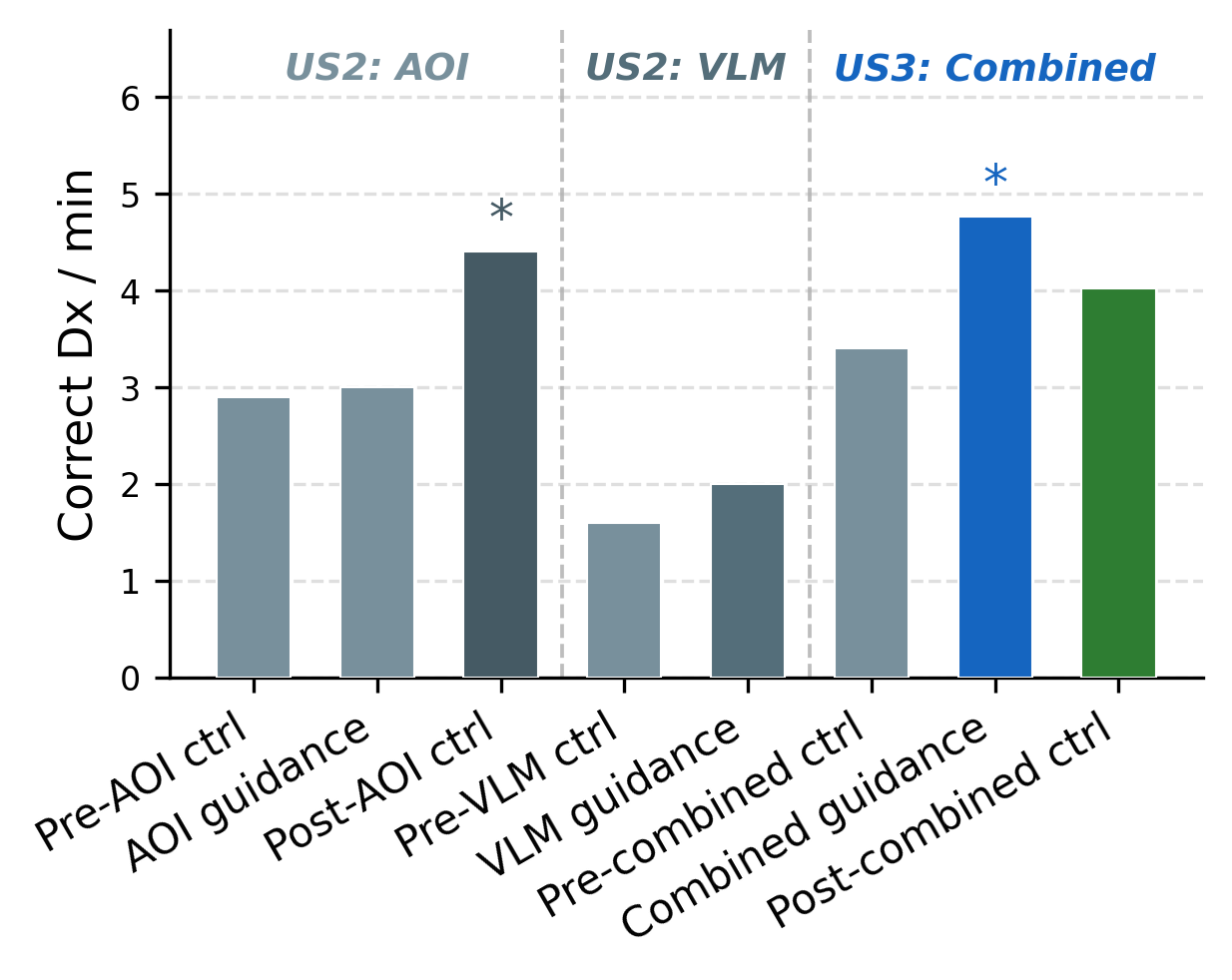}
    \caption{Cross-study Correct Dx/min: combined guidance (US3) exceeds either modality alone ($p{<}0.05$ vs.\ pre-control, Wilcoxon).}
  \end{subfigure}
  \Description{Left: three-panel boxplot showing Correct Dx/min, time per eye, and comment edit time across pre-control, combined guidance, and post-control blocks in US3. Lines connect per-participant values. Combined guidance reduces time per eye and comment edit time while increasing Correct Dx/min. Right: grouped bar chart comparing Correct Dx/min across US2 AOI guidance (3.0), US2 VLM guidance (2.0), US3 pre-control (3.40), US3 combined guidance (4.76), and US3 post-control (4.02). US3 combined guidance substantially exceeds both US2 conditions.}
  \caption{US3 efficiency results. Combined VLM+AOI guidance produces efficiency gains exceeding either modality alone (right), driven by simultaneous reductions in visual search time and documentation burden (left).}
  \label{fig:us3-results}
\end{figure*}

Combined guidance produced significant efficiency gains (\autoref{tab:all_results}): \textbf{Correct Dx/min increased from 3.40 to 4.76} ($+40\%$, 95\%~CI [$+15\%$, $+74\%$], $p=0.023$), and \textbf{comment edit time dropped from 7.2\,s to 2.4\,s} ($-67\%$, 95\%~CI [$-87\%$, $-43\%$], $p=0.023$), reflecting the VLM pre-populating comments so residents made minimal edits (\autoref{fig:us3-results}), supporting H3.2. Perceived effort also decreased significantly ($2.11 \to 1.91$, $p=0.047$). Time per eye trended lower ($26.0 \to 22.1$\,s, $-15\%$) but did not reach significance in the guidance block itself ($p=0.195$).

These gains exceeded either modality alone (\autoref{fig:us3-results}) because each modality targets a distinct workflow bottleneck: the VLM draft nearly eliminated documentation burden, while AOI heatmaps supported visual search. The combined system removed both bottlenecks at once, yielding efficiency gains neither modality could produce in isolation.

Post-guidance carryover was significant for both time per eye (19.6\,s vs.\ pre-control 26.0\,s, $p=0.016$) and time to final diagnosis (7.4\,s vs.\ 10.8\,s, $p=0.039$), supporting H3.3. The speed advantage persisted after guidance was removed, consistent with residents internalizing more efficient viewing habits during the guided block.

\majoredit{To separate genuine guidance benefit from task familiarity under the fixed block order, we fit a mixed-effects model that estimates a practice slope from the control blocks and tests the guidance effect against it (the practice-slope decomposition and the US2 mixed-effects models are detailed in \autoref{appendix:stats}). Ordering effects on diagnostic accuracy were nonsignificant, so the accuracy results are not attributable to practice; the timing carryover, by contrast, cannot be fully disentangled from familiarity under this design, which we treat as suggestive rather than confirmatory.}

Gaze–AOI divergence was stable across all three US3 blocks (all $p{>}0.6$; \autoref{tab:all_results}), in contrast to US2 AOI-only guidance where divergence increased significantly during and after guidance ($p{=}0.039$, $p{=}0.008$; \autoref{fig:divergencefigure}). Two interpretations are consistent with this pattern: the VLM text may have satisfied residents' information needs, reducing pressure to spatially re-orient gaze; or residents may have relied less on the AOI overlay in the combined condition, engaging primarily with the textual draft. Survey data showing moderate perceived reliance (3.75/7) and qualitative reports of a corroboration loop between modalities favor the first interpretation, though direct measurement of per-modality engagement is needed to distinguish the two. \majoredit{This stability is also consistent with residents' unaided gaze remaining aligned with clinically relevant regions after guidance was removed, offering suggestive (though at $n{=}8$ not statistically significant) support for the perceptual-reorientation account of the post-guidance timing carryover.}

\subsubsection{Biomarker Documentation}

\begin{figure}[t]
  \centering
  \includegraphics[width=\linewidth]{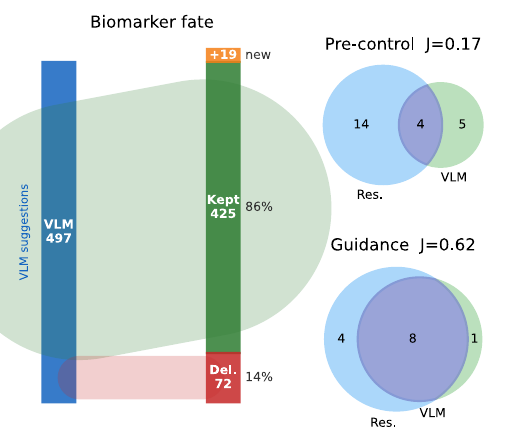}
  \Description{Three-panel figure. Left: Sankey flow diagram showing the fate of 497 VLM-suggested biomarkers in US3 (n=8): 85.5\% retained, 14.5\% deleted, and 19 new biomarkers added by residents. Center: Venn diagram for the pre-control block (Jaccard=0.17), showing low resident--VLM vocabulary overlap. Right: Venn diagram for the combined guidance block (Jaccard=0.62), showing strong vocabulary convergence while residents retained editorial agency.}
  \caption{US3 biomarker analysis: fate of VLM suggestions (left) and resident--VLM vocabulary overlap, pre-control vs.\ guidance (right). Parallel to US2 results (\autoref{fig:us2-biomarkers}).}
  \label{fig:us3-biomarkers}
\end{figure}

With combined guidance, residents documented a mean of \textbf{5.36 biomarkers per AMD eye} compared to 1.11 in the pre-guidance control ($p=0.031$, Wilcoxon signed-rank), closely matching US2 VLM-only breadth (5.8 biomarkers/eye) though marginally below the $\geq$ threshold stated in H3.4. Residents retained \textbf{85.5\%} of the 497 VLM-suggested biomarkers, meeting H3.5 ($\geq$70\% threshold; cf.\ 83.1\% in US2 VLM-only). They deleted 14.5\% of suggestions and independently contributed 19 new biomarkers not generated by the VLM, demonstrating active editorial engagement rather than passive acceptance (\autoref{fig:us3-biomarkers}). Critically, adding the AOI heatmap context did not suppress residents' editing behavior: deletion and addition rates were comparable to US2 VLM-only. Jaccard index between resident and VLM vocabulary increased from 0.17 (pre-control) to 0.62 (guidance; \autoref{fig:us3-biomarkers}), indicating strong vocabulary convergence while residents retained clinical agency.

\subsubsection{Survey and Qualitative Findings}

Post-study survey (1–7 Likert, $n{=}8$) showed residents found the guidance accurate enough to act on (mean 5.0/7) and reported applying what they learned during the post-block (4.5/7). Perceived reliance was moderate (3.75/7), consistent with using guidance as a scaffold while maintaining clinical judgment. Most residents (mean 4.0/7) preferred that guidance remain available, a stronger preference than typically observed in tool-testing contexts, suggesting perceived clinical value beyond the study setting.

Qualitatively, residents found the combined interface most valuable for uncertain or borderline cases (e.g., CNV vs.\ confluent drusen). Several described a corroboration loop between modalities: A post-graduate year 2 resident (PGY-2) noted \textit{``I'd use the heatmap and the text to correlate—if the text mentioned IRF and the heatmap highlighted a spot, that gave me confidence''}. The VLM draft was consistently appreciated as a time-saving scaffold: \textit{``The text box was incredibly helpful—I could confirm or reject entries rather than starting from scratch''} (PGY-3). Residents also noted the combined guidance helped catch secondary findings such as epiretinal membrane and posterior vitreous detachment that they would have deprioritized in unguided reading.

\majoredit{\subsection{Qualitative Analysis}
\label{sec:us3-qual}}
\majoredit{We analyzed the four open-ended prompts of the US3 post-study survey ($n{=}8$; when guidance felt helpful, when it felt unhelpful, what residents did once guidance was removed, and what single change they would make) using reflexive thematic analysis~\cite{braun2006using}. One author open-coded all responses across the four prompts, then iteratively grouped codes into candidate themes, checking each candidate against the full corpus of responses rather than isolated quotes, and consolidating codes that recurred across prompts (e.g., complaints about overlay density surfaced under both the unhelpful and change-one-thing prompts). Because this was a single-coder analysis, we treat the resulting themes as researcher-constructed interpretations reflexively grounded in the data, in the sense of Braun and Clarke's framework~\cite{braun2006using}, rather than as categories validated by inter-rater agreement. Three of the four themes that emerged mapped directly onto the three design principles we report; a fourth theme, orthogonal to the three, concerned resident control over documentation and is discussed below as a candidate fourth design consideration.}

\majoredit{\textbf{Deferrable.} Residents wanted to form an initial impression from the raw image before consulting guidance, treating it as a confirmatory check rather than a directive: \textit{``I think it was less helpful to have the AI guidance automatically on the image when I first saw it$\ldots$my first instinct was to click off the overlay and see the raw OCT image first to draw my own conclusions initially, then put the overlay back on to check the heatmap to help confirm my initial thoughts''} (P3).}

\majoredit{\textbf{Sparse.} Residents found the overlay visually overwhelming rather than selectively informative, and wanted a filterable, concise alternative: \textit{``$\ldots$if it can be more concise (eg. selectively presented) itll probably be less visually disturbing and helpful in pointing out the most important pathology$\ldots$''} (P2); \textit{``Be able to selectively highlight exactly what I'm looking for instead of everything at once. For example, if I just want to see CNV or fluid, I should be able to filter by that$\ldots$''} (P8).}

\majoredit{\textbf{Evidence-anchored.} Residents wanted each highlighted region tied to an explicit rationale, not just a location: \textit{``The hotspot look abnormal, I already know this. I want to get more information. Just telling me where to look isn't enough''} (P1). Another, asked what they still wanted once guidance was removed, requested \textit{``an explanation for why each part of the guidance was highlighted''} (P8).}

\majoredit{\textbf{Documentation scaffold (candidate fourth theme).} A theme independent of the three principles concerned residents wanting to author documentation atop an AI-provided starting point rather than accept or reject it wholesale: \textit{``I think the text annotations, would be helpful to prepopulate for my interpretation and then edit. Would also be helpful if the annotations were color-coded$\ldots$''} (P7). This theme did not map onto deferrable, sparse, or evidence-anchored; we surface it as a candidate fourth design consideration (a resident-controlled documentation scaffold) for future work rather than folding it into the existing principles.}

\begin{table*}[ht]
\centering
\small
\caption{All user study results across conditions (means). Bold = best guidance-block value per row. ``Dx''=Diagnosis. Significance vs.\ pre-guidance control (Wilcoxon, Benjamini--Hochberg): * $p{<}0.05$, ** $p{<}0.01$, $\dagger$ $p{<}0.10$; $\ddagger$ $p{<}0.05$ vs.\ guidance block. Biomarkers for wAMD eyes only. \minoredit{Gaze divergence (KL) measured in the US2 AOI condition and in US3; not in US2 VLM.} Full $p$-values in \autoref{tab:supp-pvalues}; accuracy ordering effects nonsignificant throughout (\autoref{appendix:stats}).}
\label{tab:all_results}
\begin{tabular}{l|ccc|cc|ccc}
\hline
& \multicolumn{3}{c|}{US2: AOI guidance ($n=10$)} & \multicolumn{2}{c|}{US2: VLM guidance ($n=4$)} & \multicolumn{3}{c}{US3: Combined VLM+AOI ($n=8$)} \\
\cline{2-9}
Metric & Pre-ctrl & AOI & Post-ctrl & Pre-ctrl & VLM & Pre-ctrl & Combined & Post-ctrl \\
\hline
Accuracy          & $0.927$ & $0.887$ & $0.904$                            & $0.833$ & $0.917$        & $0.833$ & $\mathbf{0.933}$\textsuperscript{$\dagger$} & $0.883$ \\
FPR               & $0.000$ & $0.007$ & $0.007$                            & $0.050$ & $0.033$        & $0.000$ & $\mathbf{0.000}$                            & $0.000$ \\
Correct Dx/min    & $2.9$   & $3.0$   & $4.4$                              & $1.6$   & $2.0$          & $3.40$  & $\mathbf{4.76}$\textsuperscript{*}          & $4.02$ \\
Time/eye (s)      & $20.0$  & $18.9$  & $14.4$\textsuperscript{*$\ddagger$}& $32.5$  & $30.1$         & $26.0$  & $22.1$                                      & $\mathbf{19.6}$\textsuperscript{*} \\
Comment edit (s)  & $4.1$   & $3.3$   & $3.1$                              & $11.3$  & $12.5$         & $7.2$   & $\mathbf{2.4}$\textsuperscript{*}           & $5.7$ \\
Time to Dx (s)    & $10.0$  & $10.8$  & $7.4$\textsuperscript{*$\ddagger$} & $14.2$  & $17.4$         & $10.8$  & $13.3$                                      & $\mathbf{7.4}$\textsuperscript{*} \\
Confidence (1--4) & $3.17$  & $3.14$  & $3.12$                             & $2.67$  & $2.83$         & $3.03$  & $3.05$                                      & $3.13$ \\
Effort (1--4)     & ---     & ---     & ---                                & $2.33$  & $2.27$         & $2.11$  & $\mathbf{1.91}$\textsuperscript{*}          & $1.98$ \\
Biomarkers/AMD eye & ---    & ---     & ---                                & $2.3$   & $\mathbf{5.8}$ & $1.11$  & $\mathbf{5.36}$\textsuperscript{*}          & $1.87$ \\
VLM retention (\%) & ---   & ---     & ---                                & ---     & $83.1$         & ---     & $\mathbf{85.5}$                             & --- \\
Gaze divergence (KL) & $3.00$ & $3.17$\textsuperscript{*} & $3.28$\textsuperscript{**} & --- & --- & $3.58$ & $3.50$ & $3.30$ \\
\hline
\end{tabular}
\Description{A table summarizing results across all three user studies (n=8 full dataset for US3). US2 AOI guidance produced a significant post-guidance carryover in time per eye; the Correct Dx/min rise was not significant. US2 VLM guidance more than doubled biomarker documentation breadth (2.3 to 5.8 per AMD eye). US3 combined guidance achieved statistically significant gains in Correct Dx/min (3.40→4.76, +40\%, p=0.023) and comment edit time (7.2→2.4s, −67\%, p=0.023), and significant post-guidance carryover for both time per eye (p=0.016) and time to diagnosis (p=0.039). Combined guidance Correct Dx/min of 4.76 substantially exceeds both AOI-only (3.0) and VLM-only (2.0), demonstrating that combined guidance exceeded either modality alone. Diagnostic accuracy was maintained throughout.}
\end{table*}

\section{Discussion}
\label{sec:discussion}


\noindent\punchline{Guidance that interrupts active sensemaking gets switched off. Clinical AI must fit the clinician's cognitive rhythm, not impose a new one.} Residents consistently adopted a \emph{read-first, check-later} rhythm, arriving at an impression before toggling guidance: \textit{``would answer first (reading the image normally), then in hindsight''} (PGY-4). When AOIs aligned with their mental model they reinforced attention; when diffuse they were switched off (\textit{``kill the noise,''} PGY-4). This explains the preserved accuracy despite the limited in-guidance efficiency gains. Clinical AI must therefore be \textbf{able to be dismissed to a later diagnostic pass by the clinician}: fast to reveal, fast to dismiss, never interrupting an active diagnostic pass.

\noindent\punchline{Clinical saliency overlays earn adoption only when they behave as sparse, precise suggestions. Comprehensive coverage undermines rather than supports expert visual search.} Heatmaps were most acceptable as \emph{suggestions}: residents asked to condense them to \textit{``2 or 3 main points of interest''} (PGY-3), with per-hotspot rationale (e.g., \textit{``IRF candidate, 0.82 confidence''}) rather than a generic saliency map. The interaction contract this implies: default to a clean image, reveal few AOIs with adaptive opacity, and tune density to experience: novices wanted upfront task scaffolding whereas seniors preferred minimal prompting.

\noindent\punchline{AI-drafted clinical text succeeds as a safety net for missed findings, not an authoritative verdict, but only when every claim is traceable to its visual evidence in the image.} Summaries worked best as a checklist: \textit{``if I saw something in the text box that I did not initially note, I would take a closer look''} (PGY-3). The primary failure mode was vagueness: ontology-specific phrasing (IRF/SRF/PED/ERM) with per-item confidence is needed, not flat prose. Three of four residents rated click-through linking between each text token and its visual source as the most helpful feature, and all preferred succinct defaults over verbose prose.

\noindent\punchline{Trustworthy clinical AI requires structural safeguards at every interaction layer: first-pass independence before any AI reveal, explicit confirmation gates for every retained item, full proposal-and-edit audit trails, and provenance marking on all auto-generated content.} A first-pass period without aids preserves independent assessment; after reveal, every retained item should require explicit confirmation with the system logging proposals and edits for accountability. Per-item confidence badges calibrate reliance (\textit{``a confidence score is most helpful,''} 3/4 residents), and very low-confidence items should collapse under ``needs review'' rather than appear as facts. Concretely, this means a linked reveal in which each biomarker token is clickable to its AOI tile beside the raw image, with explicit keep, edit, or delete decisions before anything enters the note. Expert gaze and dictation are biometric signals: any deployment must minimize retention, de-identify early, and disclose their collection in consent forms and the interface. Generalizability across scanners remains a risk; surfacing scanner context alongside an easy ``flag as incorrect'' path guards against silent drift.

\noindent\punchline{Combining AI guidance modalities is synergistic when each targets a distinct workflow bottleneck: only by removing both visual search friction and documentation burden simultaneously can the system achieve gains neither modality produces alone.} US3's combined guidance exceeded either modality alone (\autoref{tab:all_results}): the VLM draft nearly eliminated documentation burden ($-67\%$ comment edit time, $p=0.023$) while the post-guidance time-per-eye carryover ($p=0.016$) is consistent with AOIs supporting lasting perceptual reorientation, though the fixed block order means task familiarity cannot be entirely ruled out as a contributing factor. \majoredit{The same account extends to Correct Dx/min: because diagnostic accuracy is preserved across blocks rather than traded away (\autoref{tab:all_results}), the post-guidance Correct Dx/min rise seen with AOI-only guidance in US2 (2.9$\rightarrow$4.4, $+52\%$, $p{=}0.065$) is the throughput expression of the same faster unaided reading, not a change in what residents get right.} The 85.5\% biomarker retention (matching US2 VLM-only at 83.1\%) confirms AOIs did not suppress editing behavior; residents described a corroboration loop between modalities (\textit{``I'd use the heatmap and the text to correlate…''}) and wanted them linked, pointing to a design target: per-biomarker AOI tiles with clickable toggling, not global heatmaps \majoredit{(a mock-up is shown in \autoref{fig:biomarker-aoi-mockup})}.

\section{Limitations and Future Directions}

Recruiting active ophthalmology residents has been structurally con\-strained: programs typically have 12--20 trainees per institution and sessions require 30--40 minutes of clinical time. We enrolled 11 of 16 available residents in US2 (69\%) and 8 across two institutions in US3, representing substantial reach within these limited pools. Within-subjects designs and image-set-level power ($n{>}74$ per test) compensate for participant count, but detection of small individual effects and subgroup analysis await larger trials. All three studies used a fixed block sequence (pre-guidance $\to$ guidance $\to$ post-guidance), so post-guidance timing gains cannot be fully disentangled from task familiarity; accuracy ordering effects were confirmed nonsignificant (\autoref{appendix:stats}), but a crossover design would provide stronger causal evidence for carryover claims.

On the modeling side, the key limitation for the VLM is evaluation: BERTScore and MedBERTScore capture fluency rather than clinical accuracy. We will address this by bootstrapping ground-truth biomarker localizations from gaze-derived AOIs with expert ``click-to-tighten'' refinement, enabling location-aware evaluation (NSS, sAUC) and ontology-exact entity scores alongside reliance-calibration metrics.

\majoredit{\textbf{Generalizability.} 
Our study targets a binary Normal vs.\ wAMD decision on a single scanner (Zeiss Cirrus); extension to multi-class settings (wAMD, dry AMD, normal) and multisite replication will test robustness across scanners and protocols. The gaze-aligned ViT and ontology-bounded VLM are architecturally scanner-agnostic, both distill from expert gaze and dictation rather than device-specific pixel statistics, so we expect this transfer to be feasible. \majoredit{We deliberately distill from experts (US1 retina specialists) and deploy to trainees (US2--US3 residents), the population with the most to gain; attending-level clinicians were not tested, so generalization to senior clinicians remains future work.}
The methodological approach of distilling expert gaze and dictations into ViT attention alignment and ontology-bounded VLM guidance is domain-agnostic and has precedent in \majoredit{glaucoma diagnosis \cite{li2025interactively},} chest X-ray reading \cite{chen2026gaze}, and microscopy analysis in pathology \cite{kiani2020impact, ibragimov2024use}\majoredit{, though we have validated the full reading-and-documentation pipeline end-to-end only for wAMD}.
}


\section{Conclusion}

Co-Annotator distills expert gaze and dictations into two resident-facing guidance components for OCT reading in wAMD: fixation-aligned AOIs and an ontology-bounded VLM. Expert distillation sharpened both the ViT's diagnostic accuracy and the VLM's bio\-marker fidelity, and resident evaluation confirmed each modality is safe and independently beneficial: VLM guidance more than doubled biomarker documentation breadth, while AOI guidance produced a post-guidance efficiency carryover. Delivered together, they significantly increased Correct Dx/min and cut comment editing time while diagnostic accuracy was fully preserved. Neither modality could produce these gains alone. A preregistered, multisite field study comparing unaided-first vs.\ immediate guidance will test whether this design scales across scanners, protocols, and training populations.

\begin{acks}
We thank Drs.\ Royce W.S. Chen and Tony Valenzuela for their extremely helpful feedback during the development of this project and for sharing their clinical insights. We are also grateful to all the residents who participated in user studies. This research was supported in part by the Air Force Office of Scientific Research (FA9550-22-10337), the Army Research Laboratory (W911NF-19-2-0139, W911NF-19-2-0135, W911NF-21-2-0125), the U.S.\ Department of Defense (N00014-20-1-2027), and an Unrestricted Grant from Research to Prevent Blindness, Inc. awarded to the Columbia University Irving Medical Center Department of Ophthalmology.
\end{acks}

\bibliographystyle{ACM-Reference-Format}
\bibliography{main}

\appendix
\renewcommand{\thetable}{S\arabic{table}}
\setcounter{table}{0}
\renewcommand{\thefigure}{S\arabic{figure}}
\setcounter{figure}{0}

\section{\review{Dataset Curation and Preprocessing}}\label{appendix:dataset}

This section provides a detailed description of the datasets curated and preprocessed for the multi-task fine-tuning of the MedGemma VLM. Our goal was to build a specialized ophthalmic AI assistant capable of interpreting retinal OCT scans across three distinct tasks: binary diagnosis, biomarker generation, and biomarker discrimination. To achieve this, we sourced data from publicly available medical datasets, including OCT-C8 and OCT-5K, and combined them with fine-grained biomarker information extracted from physician transcripts.

\subsection{\review{Diagnosis Dataset}}

The foundation for the model's visual understanding and diagnostic capability was established using a large-scale dataset for a binary classification task.

{\setlength{\leftmargini}{1em}\setlength{\leftmarginii}{1em}
\begin{itemize}
    \item \textbf{Source:} A curated subset of the publicly available Retinal OCT-C8 dataset.
    \item \textbf{Task:} Binary classification to determine the primary pathology of an OCT scan. The model is constrained to respond with either 'Normal' or 'wet-AMD'.
    \item \textbf{Size and Composition:} The dataset comprises a total of 4,600 OCT images, balanced perfectly between the two classes (2,300 images for 'Normal' and 2,300 for 'wet-AMD').
    \item \textbf{Purpose:} This dataset's primary role is to provide a robust, domain-specific visual foundation, training the model to recognize the core features differentiating healthy retina from those affected by 
    wAMD.
\end{itemize}}

\subsection{\review{Biomarker Generation Dataset}}

A key challenge in developing specialized medical AI is the availability of high-quality, annotated data for identifying specific pathological features (biomarkers). Our biomarker dataset was designed to address this challenge for the generative identification task.

{\setlength{\leftmargini}{1em}\setlength{\leftmarginii}{1em}
\begin{itemize}
    \item \textbf{Source:} The dataset corpus was created by combining annotations from the OCT-5K dataset with a preprocessed biomarker list extracted from physician transcripts from our internal dataset.
    \item \textbf{Task:} A generative task where the model must identify all relevant pathological features from an OCT scan and output a comma-separated list of biomarkers.
    \item \textbf{Size and Composition:} The combined dataset contains a total of 573 unique samples.
    \item \textbf{Preprocessing:} To manage complexity and focus the model on the most relevant features, biomarkers with fewer than 10 occurrences in the combined dataset were excluded from the final training set. This resulted in a final vocabulary of 12 key biomarkers: "Drusen", "Photoreceptor Degeneration", "Pigment Epithelial Detachment", "Geographic Atrophy", "Choroidal Fold", "Epiretinal Membrane", "Hyperfluorescent Spots", "Intraretinal Fluid", "Posterior Vitreous Detachment", "Fluid", "Subretinal Fluid", and "Choroidal Neovascularization".
\end{itemize}}

\autoref{tab:biomarker_freq} provides a detailed frequency count of all identified biomarkers in the raw combined dataset before the final filtering step was applied. As a benchmark, we compare our biomarker distribution against the population-level reference from Kurmann et al.~\cite{Kurmann2019-zk} (\autoref{tab:biomarker-dist}); differences reflect our clinical case mix rather than extraction error.

\begin{table*}[tp]
\centering
\begin{tabular}{lccccccccccc}
\hline
 & Healthy & SRF & IRF & HF & Drusen & RPD & ERM & GA & ORA & IRC & FPED \\
\hline
Training Set (23{,}030) & 6480 & 1142 & 2947 & 5668 & 5077 & 1995 & 6139 & 1093 & 2280 & 4321 & 4766 \\
Test Set (1029)        & 165  & 65   & 48   & 178  & 376  & 153  & 140  & 62   & 200  & 54      & 359  \\
\hline
\end{tabular}
\caption{Distribution of biomarkers in Kurmann et al. \cite{Kurmann2019-zk}. \minoredit{Counts are B-scans annotated with each biomarker; a B-scan may carry several, so rows sum to more than the set size. Healthy, no pathological biomarker; SRF, subretinal fluid; IRF, intraretinal fluid; HF, hyperreflective foci; RPD, reticular pseudodrusen; ERM, epiretinal membrane; GA, geographic atrophy; ORA, outer retinal atrophy; IRC, intraretinal cysts; FPED, fibrovascular pigment epithelial detachment.}}
\label{tab:biomarker-dist}
\Description{A table listing per-biomarker B-scan counts in the Kurmann et al. reference dataset for a training set (23,030 B-scans) and a test set (1,029 B-scans) across eleven categories: Healthy, subretinal fluid, intraretinal fluid, hyperreflective foci, drusen, reticular pseudodrusen, epiretinal membrane, geographic atrophy, outer retinal atrophy, intraretinal cysts, and fibrovascular pigment epithelial detachment. A B-scan may be annotated with several biomarkers, so the counts in each row sum to more than the set size.}
\end{table*}
\begin{table}[tp]
\centering
\caption{Frequency of Biomarkers in the Combined Dataset (Prefiltering).}
\label{tab:biomarker_freq}
\Description{A two-column table listing how often each biomarker appears in the combined dataset before filtering, ordered from most to least frequent: Drusen 410, Photoreceptor Degeneration 209, Pigment Epithelial Detachment 99, Geographic Atrophy 73, Choroidal Fold 60, Posterior Vitreous Detachment 33, Hyperfluorescent Spots 32, Epiretinal Membrane 31, Intraretinal Fluid 29, Fluid 25, None 25, Subretinal Fluid 20, and Choroidal Neovascularization 12. Seven rarer biomarkers with fewer than ten occurrences (Retinal Pigment Epithelial Migration 9 down to Outer Retinal Tubulation 1) are shown in italics and were excluded from the final 12-biomarker vocabulary.}
\begin{tabular}{@{}lc@{}}
\toprule
\textbf{Biomarker}                        & \textbf{Frequency} \\ \midrule
Drusen                                    & 410                \\
Photoreceptor Degeneration                & 209                \\
Pigment Epithelial Detachment             & 99                 \\
Geographic Atrophy                        & 73                 \\
Choroidal Fold                            & 60                 \\
Posterior Vitreous Detachment             & 33                 \\
Hyperfluorescent Spots                    & 32                 \\
Epiretinal Membrane                       & 31                 \\
Intraretinal Fluid                        & 29                 \\
Fluid                                     & 25                 \\
None                                      & 25                 \\
Subretinal Fluid                          & 20                 \\
Choroidal Neovascularization              & 12                 \\
\textit{Retinal Pigment Epithelial Migration} & \textit{9}                  \\
\textit{Retinal Pigment Epithelium Atrophy}   & \textit{6}                  \\
\textit{Subretinal Hyperreflective Material}  & \textit{4}                  \\
\textit{Photoreceptor Layer Disruption}       & \textit{4}                  \\
\textit{Disciform Scar}                       & \textit{2}                  \\
\textit{Hemorrhage}                           & \textit{1}                  \\
\textit{Outer Retinal Tubulation}             & \textit{1}                  \\ \bottomrule
\end{tabular}
\end{table}

\subsection{\review{Biomarker Discrimination Dataset}}\label{appendix:distribution}

To facilitate targeted verification of specific features and improve the model's visual grounding, a VQA dataset was created.

\begin{figure*}[p]
  \centering
    \begin{subfigure}{0.46\textwidth}
        \includegraphics[width=\linewidth]{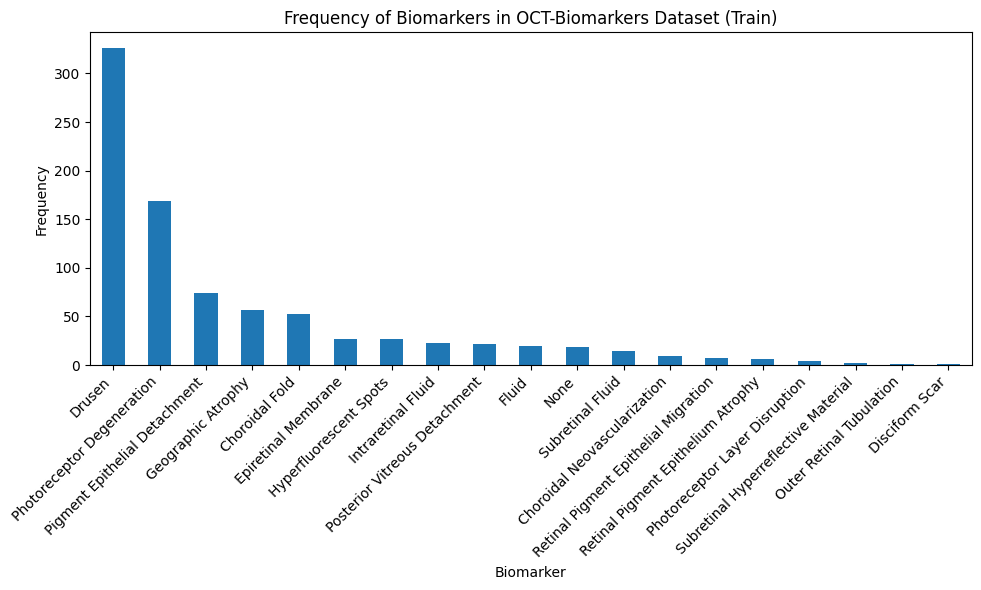}
        \caption{Training data distribution.}
        \label{fig:train_dist}
    \end{subfigure}
    \hfill
    \begin{subfigure}{0.46\textwidth}
        \includegraphics[width=\linewidth]{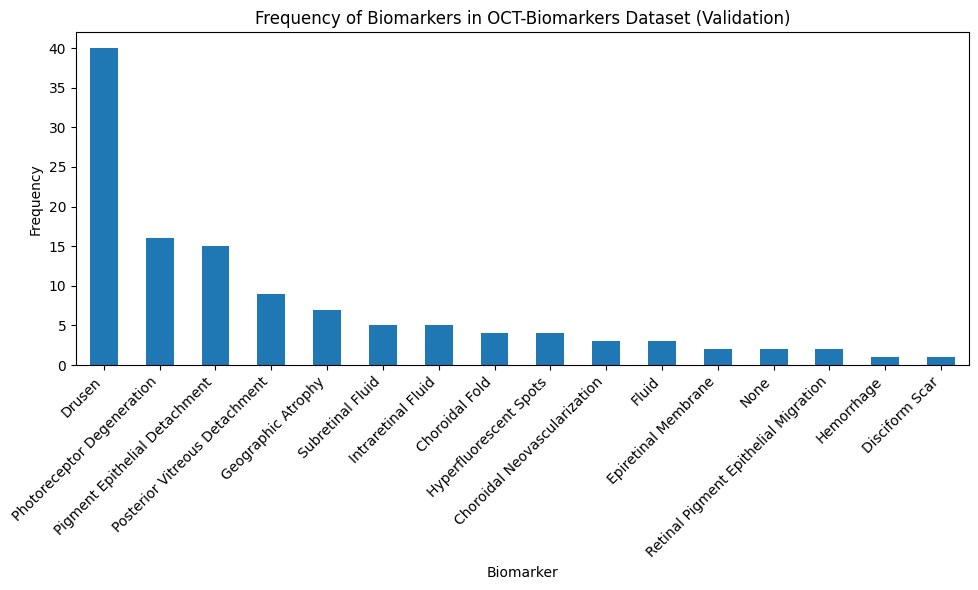}
        \caption{Validation data distribution.}
        \label{fig:val_dist}
    \end{subfigure}

    \vspace{1em}
    \begin{subfigure}{0.46\textwidth}
        \includegraphics[width=\linewidth]{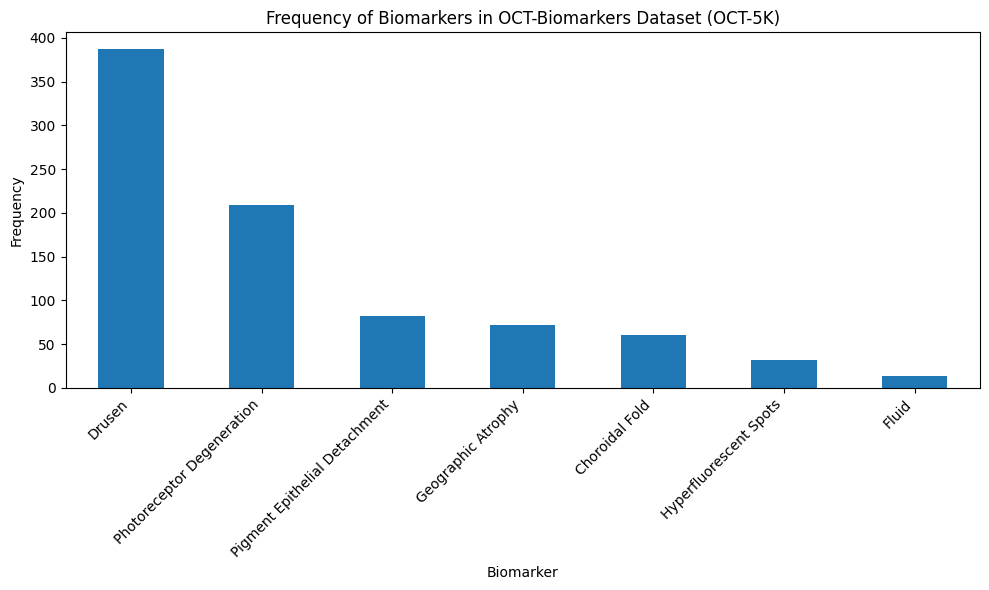}
        \caption{OCT-5K dataset distribution.}
        \label{fig:oct5k_dist}
    \end{subfigure}
    \hfill
    \begin{subfigure}{0.46\textwidth}
        \centering\includegraphics[width=\linewidth]{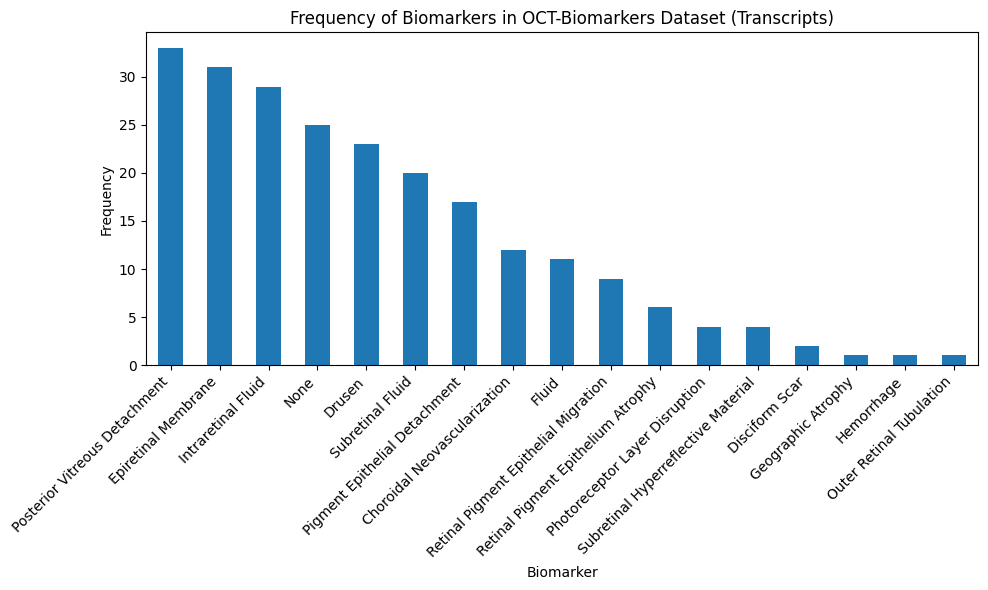}
        \caption{OCT transcript dataset distribution.}
        \label{fig:transcript_dist}
    \end{subfigure}

    \vspace{1em}
    \begin{subfigure}{0.46\textwidth}
        \centering
        \includegraphics[width=\linewidth]{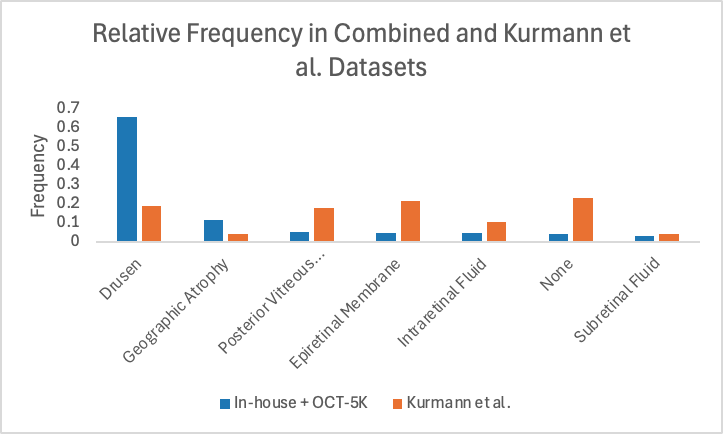}
        \caption{In-house vs.\ Kurmann et al.\ comparison.}
        \label{fig:dist-comp}
    \end{subfigure}
    \hfill
    \begin{subfigure}{0.46\textwidth}
        \centering
        \includegraphics[width=\linewidth]{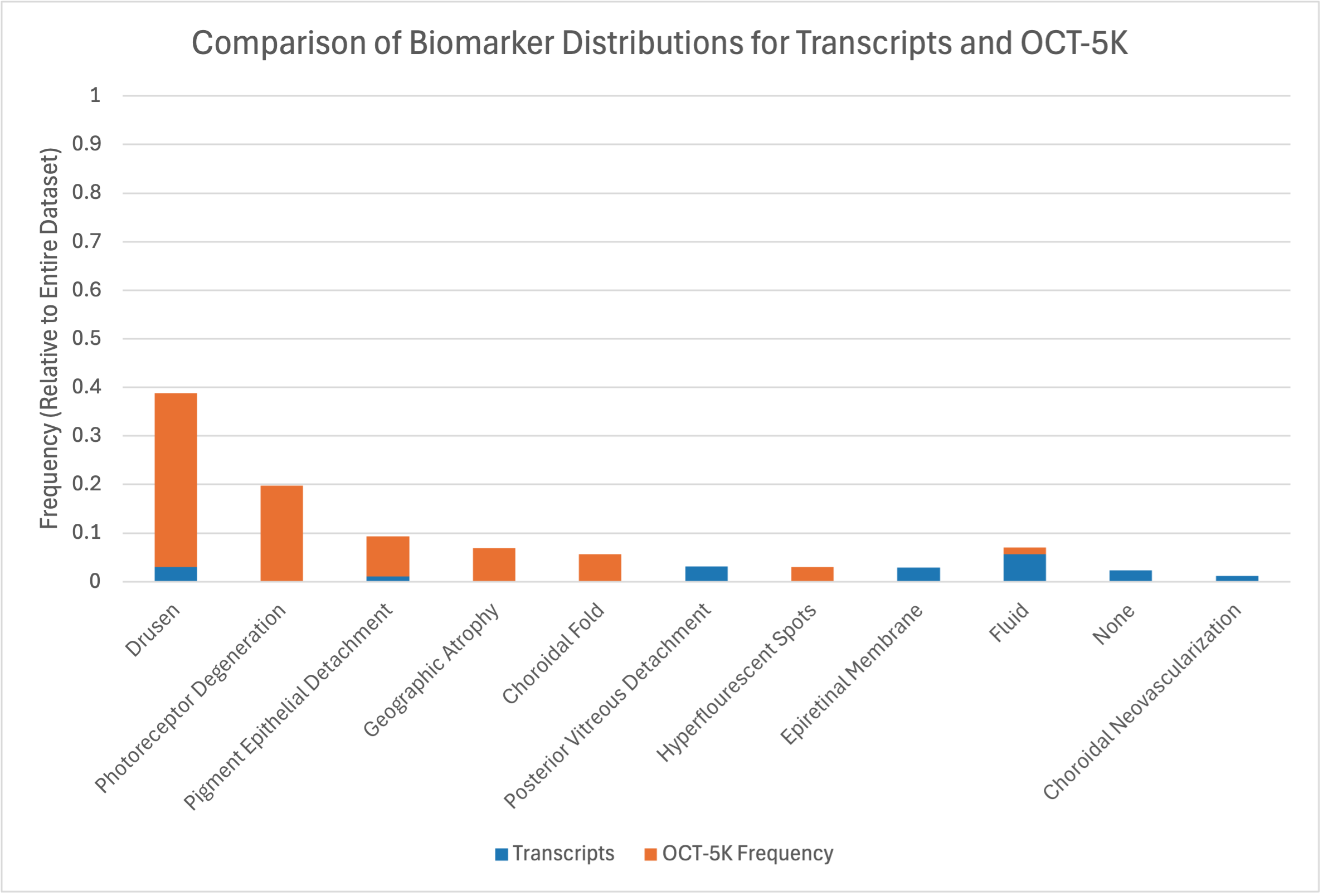}
        \caption{Transcript vs.\ OCT-5K distribution.}
    \end{subfigure}
  \caption{Biomarker dataset distribution across all data splits and sources.}
  \Description{Six bar charts showing the distribution of biomarkers across training, validation, OCT-5K, and transcript datasets, along with comparisons between datasets.}
  \label{fig:biomarker-distribs}
\end{figure*}

{\setlength{\leftmargini}{1em}\setlength{\leftmarginii}{1em}
\begin{itemize}
    \item \textbf{Source:} The biomarker discrimination dataset was dynamically generated from the curated biomarker generation Dataset.
    \item \textbf{Size and Composition:} The dataset comprises a total of 1664 unique samples.
    \item \textbf{Task:} A discriminative task where the model is presented with a direct question of the form, "Is [BIOMARKER\_NAME] present in this OCT scan?" and must respond with a definitive 'Yes' or 'No'.
    \item \textbf{Generation Strategy:} An initial, exhaustive generation process revealed that an uncurated discriminative dataset would be overwhelmingly populated with "No" answers, potentially teaching the model a trivial strategy of always predicting "No". To circumvent this, a balancing strategy was employed. For each image, an equal number of "Yes" instances (for biomarkers present) and "No" instances (randomly selected from biomarkers that were absent) were generated. This approach forced the model to learn the actual visual features corresponding to each biomarker rather than relying on statistical priors.
    \item \textbf{Representativeness} Generated biomarkers were diverse and representative in identity to those known in wAMD \cite{metrangolo_oct_2021, hanson_optical_2023}. Though some differ appreciably (\autoref{fig:biomarker-distribs}) in frequency compared to established datasets, this is likely due to our different case mix rather than detection error. 
\end{itemize}}

\section{Biomarker VLM Details}\label{appendix:vlm-training}
The VLM fine-tuning was conducted in two stages. The detailed configurations for each stage are provided in \autoref{tab:training_configs}. \minoredit{Stage two, which adds the biomarker-generation task, uses a balanced sampler that enforces uniform task mixing per batch across all three tasks (diagnosis, biomarker discrimination, biomarker generation). This prevents catastrophic forgetting of the diagnosis and discrimination objectives learned in stage one while forcing the model to map visual features to ontology-bounded biomarker labels for the new generative task.}

\subsection{System Prompt}\label{appendix:vlm-prompt}

The following system prompt is prepended to all three task queries to enforce ontology-bounded, format-constrained responses:

\begin{quote}\small
\textbf{System:} You are an expert ophthalmic AI assistant. Analyze the provided retinal OCT scan. Respond concisely and accurately, sticking strictly to the requested format without additional explanation or conversational text.

For diagnostic queries, provide only the diagnosis (\texttt{wet-AMD} or \texttt{Normal}). For biomarker identification, provide only a comma-separated list of findings (e.g., \textit{Drusen, Subretinal fluid}) or \texttt{None} if no biomarkers are present. For direct questions, answer only with \texttt{Yes} or \texttt{No}. Do not add introductory phrases, explanations, or disclaimers.
\end{quote}

\begin{table*}[t]
  \centering
  \caption{Training configurations for the VLM fine-tuning.}
  \label{tab:training_configs}
  \Description{A table of VLM fine-tuning configurations comparing Stage One and Stage Two. Both stages use the unsloth MedGemma-4b-it 4-bit model with identical LoRA settings (rank 32, alpha 64, dropout 0.05) and trainer settings (train batch size 56, eval batch size 224, gradient accumulation 1, warmup ratio 0.1). Stage One trains for 3 epochs at learning rate 5e-6; Stage Two trains for 8 epochs at learning rate 1e-6.}
  \begin{tabular}{@{}lll@{}}
    \toprule
    \textbf{Parameter} & \textbf{Stage One Configs} & \textbf{Stage Two Configs} \\
    \midrule
    \multicolumn{3}{l}{\textit{Model Configuration}} \\
    \quad Model Name & unsloth/medgemma-4b-it-unsloth-bnb-4bit & unsloth/medgemma-4b-it-unsloth-bnb-4bit \\
    \midrule
    \multicolumn{3}{l}{\textit{LoRA Configuration}} \\
    \quad r & 32 & 32 \\
    \quad alpha & 64 & 64 \\
    \quad dropout & 0.05 & 0.05 \\
    \midrule
    \multicolumn{3}{l}{\textit{Trainer Configuration}} \\
    \quad Num Epochs & 3 & 8 \\
    \quad Train Batch Size & 56 & 56 \\
    \quad Eval Batch Size & 224 & 224 \\
    \quad Grad Accum Steps & 1 & 1 \\
    \quad Warmup Ratio & 0.1 & 0.1 \\
    \quad Learning Rate & 5e-6 & 1e-6 \\
    \bottomrule
  \end{tabular}
\end{table*}

\review{
\subsection{Per-Biomarker Evaluation Performance}}
\label{appendix:per_biomarker_eval}

To address the need for clinical specificity beyond semantic similarity metrics (e.g., BERTScore), we conducted a granular, ontology-exact evaluation of the VLM. We calculated F1, Precision, and Recall scores for individual biomarkers across two distinct tasks: \textit{Biomarker Generation} (where the model spontaneously lists findings) and \textit{Biomarker Discrimination} (where the model answers "Yes/No" to specific queries, e.g., "Is Drusen present?").

The evaluation was conducted on a validation set of 58 samples. As shown in \autoref{tab:per_biomarker_metrics}, the model demonstrates strong performance on common pathologies but exhibits performance degradation on rare classes, highlighting the impact of data imbalance in the training corpus. The model performs better when explicitly queried about specific findings (discrimination) compared to when generating a list of findings. 

\begin{table*}[tp]
\centering
\caption{Per-Biomarker Performance Metrics. Comparison of the Generative (List) task vs. the Discriminative (VQA) task. $N$ represents the number of support examples in the validation set. Evaluation is ontology-exact.}
\label{tab:per_biomarker_metrics}
\Description{A table of per-biomarker performance (F1, Precision, Recall, and support count N) on a 58-sample validation set, comparing the generative list task against the discriminative yes/no task for ten biomarkers. Common biomarkers score highest, for example Drusen (generation F1 0.87, discrimination F1 0.98) and Photoreceptor Disruption (0.67 and 0.88); several rare biomarkers score F1 0.00. The model generally performs better on the discrimination task than the generation task, for example Intraretinal Fluid rises from generation F1 0.50 to discrimination F1 1.00.}
\small
\begin{tabular}{lcccccccc}
\toprule
& \multicolumn{4}{c}{\textbf{Biomarker Generation Task}} & \multicolumn{4}{c}{\textbf{Biomarker Discrimination Task}} \\
\cmidrule(lr){2-5} \cmidrule(lr){6-9}
\textbf{Biomarker} & \textbf{F1} & \textbf{Precision} & \textbf{Recall} & \textbf{$N$} & \textbf{F1} & \textbf{Precision} & \textbf{Recall} & \textbf{$N$} \\
\midrule
Drusen & 0.87 & 0.83 & 0.91 & 44 & 0.98 & 0.96 & 1.00 & 46 \\
Photoreceptor Disruption & 0.67 & 0.58 & 0.79 & 24 & 0.88 & 0.82 & 0.96 & 29 \\
Pigment Epithelial Detachment & 0.40 & 0.33 & 0.50 & 10 & 0.38 & 0.50 & 0.30 & 18 \\
Intraretinal Fluid & 0.50 & 0.33 & 1.00 & 1 & 1.00 & 1.00 & 1.00 & 16 \\
Epiretinal Membrane & 0.00 & 0.00 & 0.00 & 2 & 0.67 & 1.00 & 0.50 & 12 \\
Choroidal Fold & 0.50 & 0.40 & 0.67 & 3 & 0.00 & 0.00 & 0.00 & 15 \\
Posterior Vitreous Detachment & 0.29 & 0.20 & 0.50 & 2 & 0.00 & 0.00 & 0.00 & 17 \\
Geographic Atrophy & 0.00 & 0.00 & 0.00 & 9 & 0.20 & 1.00 & 0.11 & 14 \\
Fluid (Generic) & 0.29 & 0.33 & 0.25 & 4 & 0.00 & 0.00 & 0.00 & 11 \\
Subretinal Fluid & 0.00 & 0.00 & 0.00 & 1 & 0.00 & 0.00 & 0.00 & 13 \\
\bottomrule
\end{tabular}
\end{table*}

\review{\subsection{Analysis of Task Performance}}
We observed two distinct trends in the model's behavior:

\paragraph{Task Complexity (Discrimination vs. Generation)}
The model generally achieved higher performance on the \textit{Biomarker Discrimination} task compared to \textit{Biomarker Generation}. For example, detection of Intraretinal Fluid improved from an F1 of 0.50 in the generative setting to 1.00 in the discriminative setting. This performance gap is expected, as the binary classification nature of the discrimination task is inherently simpler than the open-ended generation required to list all present biomarkers. Additionally, the Discrimination dataset is larger and balanced with adversarial "No" examples, providing stronger supervision for specific features.

\paragraph{Frequency Bias}
Biomarkers with high prevalence in the training set, such as Drusen ($N=46$ in discrimination test) and Photoreceptor Disruption, achieved high F1 scores across both tasks (Drusen Discrimination F1: 0.98). Conversely, low-frequency biomarkers such as Subretinal Fluid and Epiretinal Membrane suffered from low detection rates. Note that biomarkers with zero support samples in the pilot test set (e.g., CNV, Hemorrhage) were excluded from this analysis. Future work will necessitate a larger, stratified test set to ensure robust evaluation of these rare pathologies.

\review{\section{Gaze-Alignment Vision Transformer (ViT) Details}}\label{appendix:vit-training}

This section provides the technical specifications for reproducing the gaze-aligned ViT training described in \autoref{sec: vit}. Architecture, training hyperparameters, loss configuration, and data augmentation are summarized in \autoref{tab:vit-hyperparams}.

\begin{table}[t]
\centering
\small
\caption{ViT performance vs.\ alignment weight $\alpha$ (averaged across five folds). Val.\ = Validation. The best trade-off is at $\alpha=0.05$.}
\label{tab:vit results}
\begin{tabular}{lcc}
\toprule
\textbf{$\alpha$} & \textbf{Val.\ Accuracy} & \textbf{Val.\ F1} \\
\midrule
0 (no attention-alignment) & 0.8607 & 0.8380 \\
0.01 & 0.8403 & 0.8073 \\
\textbf{0.05} & \textbf{0.8837} & \textbf{0.8628} \\
0.1  & 0.8347 & 0.8140 \\
0.2  & 0.8687 & 0.8500 \\
0.3  & 0.8630 & 0.8429 \\
\bottomrule
\end{tabular}
\Description{A table showing vision transformer diagnostic accuracy by varied alignment weights, alpha. Validation accuracy was maximized at alpha = 0.05 (accuracy 0.8837, F1 0.8628).}
\end{table}

\review{\subsection{Fixation Processing}}

\majoredit{For each image, fixation events were detected from the raw gaze stream via velocity-thresholding~\cite{salvucci2000identifying} and converted to a fixation-density heatmap by kernel density estimation. Heatmaps were normalized to unit mass and then downsampled to the $32\times32$ ViT patch grid, yielding the rasterized attention target $A^*$ used in the alignment loss. Because $A^*$ accumulates over the full diagnostic review, it links gaze to the holistic diagnosis rather than to individual biomarker tokens.}

\review{\subsection{Attention Rollout Algorithm}}

\review{Attention rollout computes the effective attention from input patches to the final layer by recursively multiplying attention matrices across all transformer layers.}

\paragraph{Given:}
\begin{itemize}
    \item $A^{(\ell)} \in \mathbb{R}^{N \times N}$: average attention matrix at layer $\ell$ (averaged over all 12 heads),
    \item $N = 1025$ tokens (1 [CLS] token + $32 \times 32 = 1024$ patches),
    \item $L = 12$ transformer layers.
\end{itemize}

\paragraph{Algorithm:}

\begin{enumerate}
    \item \textbf{Initialize:}
    \[
        \tilde{A}^{(0)} = I.
    \]

    \item \textbf{Recursive aggregation} for each layer $\ell = 1, 2, \ldots, 12$:
    \[
        \tilde{A}^{(\ell)} = \tilde{A}^{(\ell - 1)} \cdot A^{(\ell)}.
    \]

    \item \textbf{Extract attention} from the [CLS] token to image patches:
    \[
        \hat{A} = \tilde{A}^{(L)}[0,\, 1:]
    \]
    (i.e., the first row of $\tilde{A}^{(L)}$ excluding the [CLS] position).

    \item \textbf{Reshape and normalize:}
    \begin{align*}
        \hat{A}_{\text{grid}} &= \mathrm{reshape}(\hat{A},\, (32,32)), \\
        \hat{A}^{(L)}(x) &= \frac{\hat{A}_{\text{grid}}}{\sum_{i,j} \hat{A}_{\text{grid}}[i,j]}.
    \end{align*}
\end{enumerate}

This yields a normalized $32 \times 32$ probability distribution $\hat{A}^{(L)}(x)$ over image patches, directly comparable to the expert fixation-density target $A^*$.

\begin{table}[tp]
\centering
\caption{ViT gaze-alignment training hyperparameters and configuration.}
\label{tab:vit-hyperparams}
\Description{A table of the gaze-aligned Vision Transformer training configuration. Architecture: ViT-Base/16 with 12 layers, hidden dimension 768, 12 attention heads, MLP dimension 3072, 16 by 16 patches, 512 by 512 input, a 32 by 32 patch grid, and 1025 tokens; initialized from ImageNet-21k weights. Training: AdamW optimizer, base learning rate 3e-4 (minimum 1e-6), betas 0.9 and 0.999, weight decay 0.05, gradient clipping at max-norm 1.0, batch size 32, 100 epochs with 10 warmup epochs and cosine-annealing schedule. Loss: binary cross-entropy classification plus a cross-entropy attention-alignment term, with alpha values 0, 0.01, 0.05, 0.1, 0.2, and 0.3 tested and 0.05 selected. The table also lists data-augmentation settings.}
\small
\begin{tabular}{l l}
\toprule
\textbf{Model Architecture} & \\
ViT variant & Base/16 \\
Number of layers & 12 \\
Hidden dimension & 768 \\
Number of heads & 12 \\
MLP dimension & 3072 \\
Patch size & $16 \times 16$ \\
Input resolution & $512 \times 512$ \\
Patch grid size & $32 \times 32$ \\
Total tokens & 1025 (1 [CLS] + 1024 patches) \\
\midrule
\textbf{Initialization} & \\
Pretrained weights & ImageNet-21k \\
\midrule
\textbf{Training Configuration} & \\
Optimizer & AdamW \\
Base learning rate & $3 \times 10^{-4}$ \\
Minimum learning rate & $1 \times 10^{-6}$ \\
$\beta_1, \beta_2$ & 0.9, 0.999 \\
Weight decay & 0.05 \\
Gradient clipping & max\_norm = 1.0 \\
Batch size & 32 \\
Total epochs & 100 \\
Warmup epochs & 10 \\
LR schedule & Cosine annealing \\
\midrule
\textbf{Loss Function} & \\
Classification loss & Binary cross-entropy (BCE) \\
Alignment loss & Cross-entropy between distributions \\
$\alpha$ values tested & \{0, 0.01, 0.05, 0.1, 0.2, 0.3\} \\
Selected $\alpha$ & 0.05 \\
\midrule
\textbf{Data Augmentation} & \\
Horizontal flip & $p=0.5$ \\
Color jitter & brightness=0.2, contrast=0.2 \\
Random rotation & $\pm 10^\circ$ \\
Random crop & scale=[0.9, 1.0] \\
\bottomrule
\end{tabular}
\end{table}

\review{\subsection{Cross-Entropy Loss for Probability Distributions}}

The alignment loss compares two discrete probability maps over the $32 \times 32$ patch grid:
\begin{equation*}
    \mathcal{L}_{\mathrm{align}}
     \;=\;
     -\sum_{i=1}^{32} \sum_{j=1}^{32}
     A^*[i,j]\,
     \log\!\left(\hat{A}^{(L)}[i,j] + \epsilon \right),
\end{equation*}
where:
\begin{itemize}
    \item $A^*[i,j]$ is the expert fixation density, normalized such that $\sum_{i,j} A^*[i,j]=1$.
    \item $\hat{A}^{(L)}[i,j]$ is the model attention from attention rollout.
    \item $\epsilon = 10^{-8}$ ensures numerical stability.
\end{itemize}

\review{\subsection{Training Procedure}}
For each training image $x$ with label $y$ and expert target $A^*$:

\begin{enumerate}
    \item \textbf{Forward pass:}
    \[
        \hat{y} = \mathrm{ViT}(x), \qquad
        \hat{A}^{(L)} = \mathrm{AttentionRollout}(\mathrm{ViT},\, x).
    \]

    \item \textbf{Compute losses:}
    \begin{align*}
        \mathcal{L}_{\mathrm{cls}} &= \mathrm{BCE}(\hat{y}, y), \quad
        \mathcal{L}_{\mathrm{align}} = \mathrm{CrossEntropy}\big(\hat{A}^{(L)}, A^*\big), \\
        \mathcal{L}_{\mathrm{total}} &= \mathcal{L}_{\mathrm{cls}} + \alpha\,\mathcal{L}_{\mathrm{align}}.
    \end{align*}

    \item \textbf{Backward pass and optimization:}
    \begin{itemize}
        \item Compute gradients via backpropagation.
        \item Clip gradients with max\_norm = 1.0.
        \item Update weights using AdamW.
    \end{itemize}
\end{enumerate}

\paragraph{Learning rate schedule:}
\begin{itemize}
    \item Linear warmup (epochs 1--10) from 0 to $3\times10^{-4}$.
    \item Cosine annealing (epochs 11--100) down to $1\times10^{-6}$.
\end{itemize}

\paragraph{Model selection:}
The checkpoint with highest validation accuracy across the 5 folds is retained. Final results are averaged across folds.

\review{\subsection{Implementation Notes}}

{\setlength{\leftmargini}{1em}\setlength{\leftmarginii}{1em}
\begin{itemize}
    \item \textbf{Attention extraction overhead:} Attention rollout adds $\sim$15\% computational overhead.
    \item \textbf{Patient-level Cross-validation:} Data split at the patient level prevents leakage, since each patient contributes five OCT images.
    \item \textbf{Baseline:} $\alpha=0$ corresponds to a standard ViT trained solely with classification loss.
    \item \textbf{Mixed precision:} Automatic Mixed Precision (AMP) reduces memory usage and improves training speed without affecting final accuracy.
    \item \textbf{Attention normalization:} Rollout-generated attention maps are always normalized to sum to 1 before computing loss.
\end{itemize}}

\section{Inter-Participant Agreement}

The results of calculation of Cohen's $\kappa$ across the image sets rated by two different participants in each user study set are presented in \autoref{tab:aoi_kappa}. 

\begin{table}[t]
    \centering
    \begin{tabular}{c|cccccc}
         & $\kappa$ & 97.5\% CI & $N$  \\
         \hline
         post-guidance control & 0.48 & (0.15, 0.81) & 18 \\
         AOI-guidance & 0.66 & (0.36,0.96) & 24 \\
    \end{tabular}
    \caption{$\kappa$ values and confidence intervals for inter-reader agreement across user study conditions.}
    \label{tab:aoi_kappa}
    \Description{A table of Cohen's kappa inter-reader agreement across US2 conditions: post-guidance control has kappa 0.48 (97.5 percent confidence interval 0.15 to 0.81, N=18), and AOI guidance has kappa 0.66 (confidence interval 0.36 to 0.96, N=24).}
\end{table}

\subsection{US2 VLM Biomarker Documentation Analysis}

\begin{figure*}[t]
  \centering
    \begin{subfigure}{0.32\textwidth}
        \includegraphics[width=\linewidth]{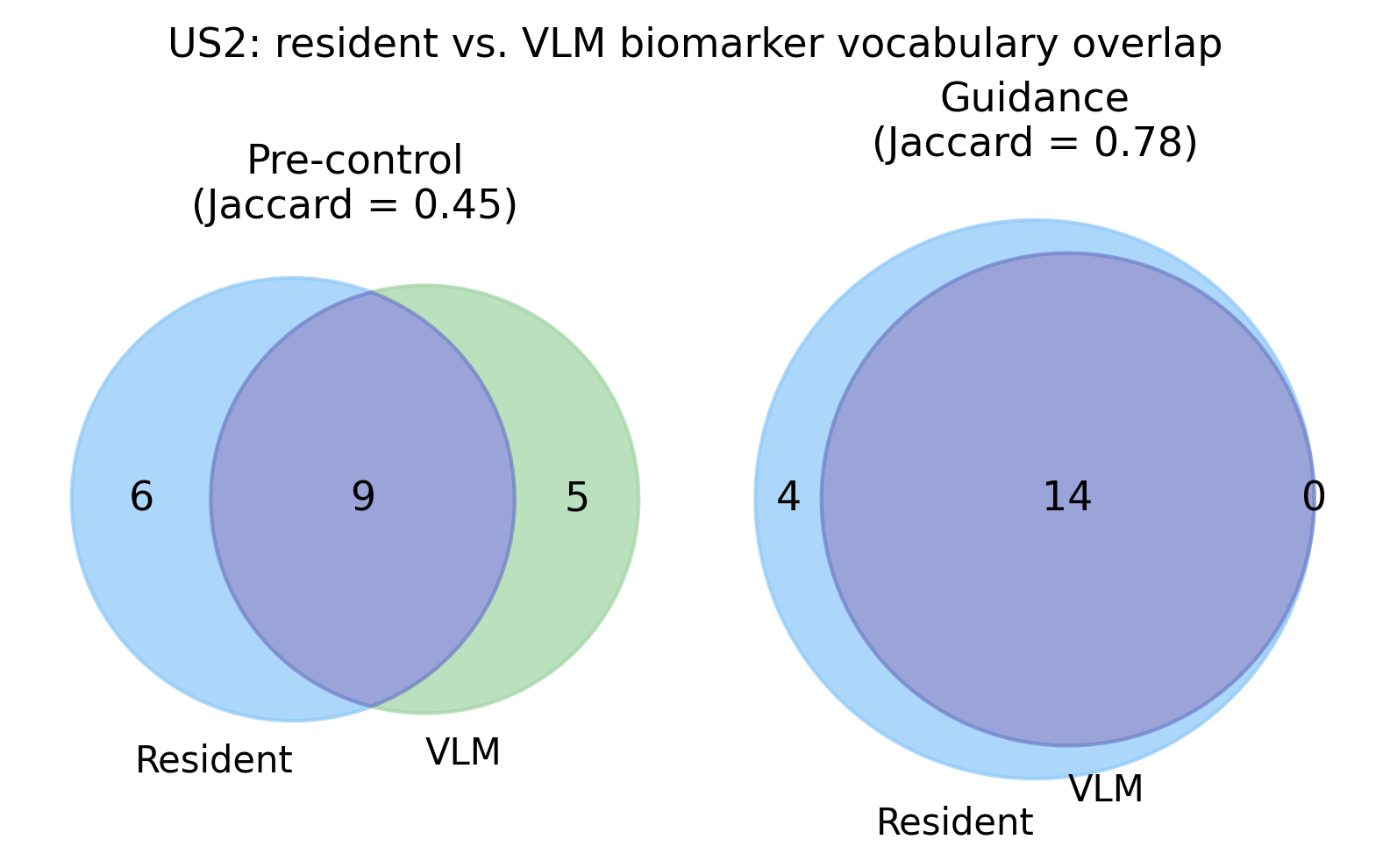}
        \caption{Biomarker vocabulary overlap (control vs.\ guidance).}
        \label{fig:VLMvenn}
    \end{subfigure}
    \hfill
    \begin{subfigure}{0.32\textwidth}
        \includegraphics[width=\linewidth]{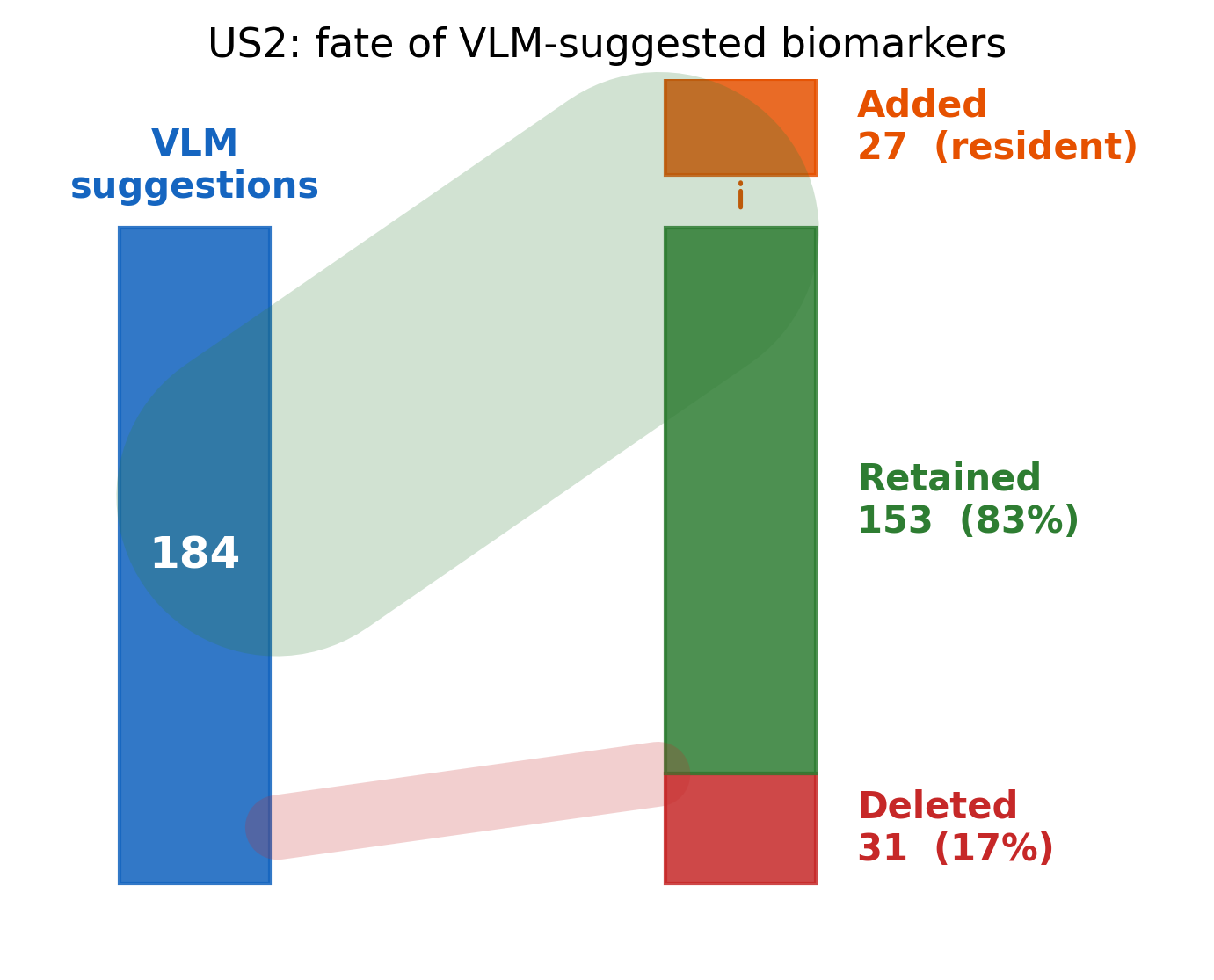}
        \caption{Fate of VLM-suggested biomarkers after resident review.}
        \label{fig:biomarkersankey}
    \end{subfigure}
    \hfill
    \begin{subfigure}{0.32\textwidth}
        \includegraphics[width=\linewidth]{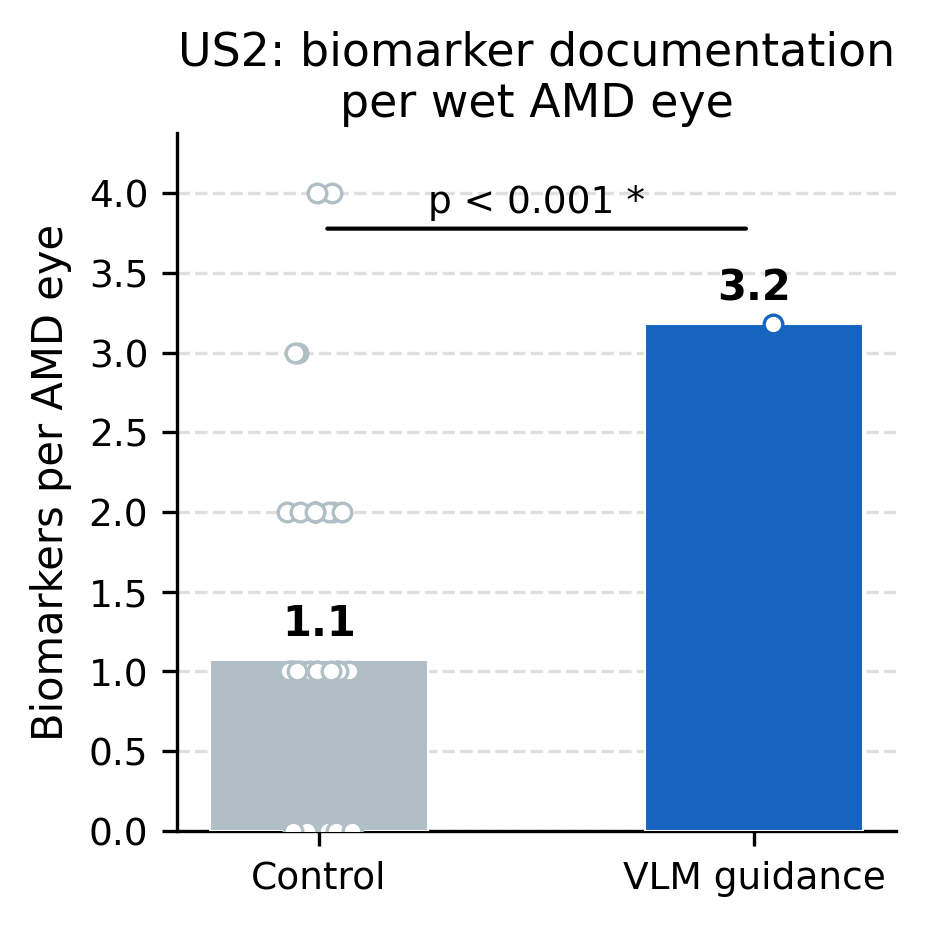}
        \caption{Biomarkers per AMD eye: control vs.\ guidance.}
        \label{fig:biomarkerdistrib}
    \end{subfigure}
  \Description{A multi-part figure detailing the biomarker analysis from the US2 VLM experiment. (a) Two Venn diagrams showing resident vs.\ VLM vocabulary overlap in the control block (Jaccard 0.45) and guided block (Jaccard 0.78). (b) Sankey diagram tracking fates of 184 VLM-suggested biomarkers: 83.1\% retained, 31 deleted, 3 negated, 27 added by residents. (c) Distributions of biomarkers documented per block, showing greater breadth in the VLM guidance condition.}
  \caption{US2 VLM guidance: biomarker vocabulary and documentation analysis. Parallel to US3 results (\autoref{fig:us3-biomarkers}); retention rate (83.1\%) is consistent across both studies.}
  \label{fig:us2-biomarkers}
\end{figure*}

\autoref{fig:us2-biomarkers} details the biomarker vocabulary and documentation outcomes from the US2 VLM guidance condition, parallel to the US3 analysis in the main paper (\autoref{fig:us3-biomarkers}). Residents retained 83.1\% of VLM-suggested biomarkers, consistent with the 85.5\% retention observed in US3.

\review{\section{Hypothesis Tests and Mixed-Effects Models for US2 and US3}}
\label{appendix:stats}

\autoref{tab:supp-pvalues} reports raw and Benjamini--Hochberg adjusted $p$-values for all primary outcome measures in US2. Mixed-effects model estimates for the ordering effect decomposition are in \autoref{tab:mixed-correct} (accuracy) and \autoref{tab:mixed-time} (response time).

\begin{table*}[ht]
\centering
\begin{tabular}{c|ccc|c}
    \toprule
    && AOI & & VLM\\
  \hline
 &pre vs. guided & guided vs. post & pre vs. post& pre vs. guided \\ 
  \hline
 accuracy & 0.321 (0.936) & 0.702 (0.936) & 0.527 (0.936) & 0.269 (0.936) \\ 
   FPR & 1 (1) & 1 (1) & 0.474 (0.936) & 1 (1) \\ 
   FNR & 0.42 (0.936) & 0.692 (0.936) & 0.668 (0.936) & 0.322 (0.936) \\ 
   Time per eye & 0.608 (0.751) & $1.42 \times 10^{-4}$ ($7.44 \times 10^{-4}$) & $6.38 \times 10^{-6}$ ($1.34 \times 10^{-4}$) & 0.198 (0.519) \\ 
   Correct dx per min & 0.912 (0.957) & 0.113 (0.395) & 0.065 (0.274) & 0.486 (0.751) \\ 
   Time to final Dx & 0.296 (0.622) & $1.48 \times 10^{-5}$ ($1.55 \times 10^{-4}$) & $1.31 \times 10^{-4}$ ($7.44 \times 10^{-4}$) & 0.477 (0.751) \\ 
   Confidence & 0.71 (0.785) & 0.968 (0.968) & 0.678 (0.785) & 0.25 (0.583) \\ 
   comment edit time & 0.359 (0.686) & 0.523 (0.751) & 0.141 (0.424) & 0.542 (0.751) \\ 
   Effort &  &  &  & 0.582 (0.751) \\ 
   \bottomrule
\end{tabular}
\caption{\textit{P}-values and Benjamini--Hochberg adjusted \textit{p}-values (in parentheses) for measured quantities by experiment blocks in AOI and VLM conditions in \textbf{US2}. Values for accuracy, FPR, and FNR were generated by Fisher's Exact test. The Wilcoxon signed-rank test was used for all other quantities.}
    \label{tab:supp-pvalues}
    \Description{A table of raw and Benjamini-Hochberg-adjusted p-values (adjusted values in parentheses) for US2 outcome measures across block comparisons, with three columns for the AOI condition (pre versus guided, guided versus post, pre versus post) and one for the VLM condition (pre versus guided). Rows are accuracy, false-positive rate, false-negative rate, time per eye, correct diagnoses per minute, time to final diagnosis, confidence, comment edit time, and effort. Accuracy, FPR, and FNR used Fisher's exact test; other measures used the Wilcoxon signed-rank test. Timing effects are strongest: in the AOI condition, time per eye and time to final diagnosis are significant for the guided-versus-post and pre-versus-post comparisons (adjusted p below 0.001), while accuracy measures are non-significant throughout.}
\end{table*}


\review{\section{Mixed Effects Modeling in US2}}

The results of mixed effects modeling for the AOI and VLM conditions on diagnostic correctness (whether participant diagnosis agrees with ground truth diagnosis) and time-to-diagnosis are presented in \autoref{tab:mixed-correct}--\autoref{tab:mixed-time}. We allowed for random by-participant effects on the diagnostic correctness while controlling for ground truth classification (normal vs wAMD) and image position in the experimental block (to control for possible fatigue effects). The latter two were considered fixed effects to ensure model convergence. Modeling was performed using the `lmer` package in R.

\begin{table*}[tp]
\centering
\begin{tabular}{llccc}
\textbf{AOI experiment}&Variable & Log-Odds & Std. Error & $p$ \\
  \hline
  \textit{Fixed effects}&Intercept & 4.098 & 0.997 & $3.98 \times 10^{-5}$ \\ 
&  AOI guidance & 0.647 & 0.887 & 0.466\\ 
&  post-guidance control & 1.260 & 0.928 &0.174 \\ 
&  GT = wAMD & $-2.979$ & 0.727 &  $4.21 \times 10^{-5}$ \\ 
&  within-block position & 0.175 & 0.087 & 0.044 \\ 
&  AOI guidance:within-block position & $-0.162$ & 0.109 &  0.137 \\ 
  &post-guidance control:within-block position & $-0.191$ & 0.114 &  0.093 \\ 
   \hline
  &Group & Name & Variance & Std. Dev. \\ 
 \cline{2-5}
 
 \textit{Random effects} & Participant & Intercept & 1.337 & 1.156\\
\hline
&&AIC & BIC & log(likelihood) \\
\cline{2-5}
\textit{Model fit}&&232.7& 265.3&$-108.3$\\
&&&&\\
\textbf{VLM experiment}&Variable & Log-Odds & Std. Error & $p$ \\
\hline
  \textit{Fixed effects}&Intercept & 2.565 & 0.899 & $4.35 \times 10^{-3}$  \\ 
&  VLM guidance & 0.441 & 1.349 &  0.744 \\ 
  &GT = wAMD & $-0.337$ & 0.616 &  0.584 \\ 
  &within-block position & $-0.087$ & 0.083 &  0.296 \\ 
  &VLM guidance control:within-block position & 0.043 & 0.137 &  0.754 \\ 
 \hline
 & Group & Name & Variance & Std. Dev. \\ 
 \cline{2-5}
 \textit{Random effects} &Participant&Intercept&0.0343&0.185\\
 \hline
 &&AIC & BIC & log(likelihood) \\
\cline{2-5}
\textit{Model fit}&&98.7& 115.4&$-43.4$\\
 
\end{tabular}
    \caption{Mixed effects modeling of diagnostic correctness in each experimental condition of \textbf{US2}. ``GT" = Ground Truth, ``:" = interaction between, AIC = Akaike Information Criterion, BIC = Bayesian Information Criterion.}
    \label{tab:mixed-correct}
    \Description{A table reporting mixed-effects logistic models of diagnostic correctness in US2 for the AOI and VLM experiments. For each experiment it lists fixed-effect log-odds, standard errors, and p-values for the intercept, the guidance and post-guidance-control blocks, ground-truth wAMD, within-block position, and their interactions; a by-participant random intercept; and model-fit statistics (AIC, BIC, log-likelihood). Guidance and post-guidance-control block effects on correctness are non-significant (for example, AOI guidance log-odds 0.647, p=0.466), while ground-truth wAMD and within-block position are significant predictors in the AOI model.}
\end{table*}

\begin{table*}[tp]
\centering
\begin{tabular}{llcccc}
 \textbf{AOI experiment}& Variable & Coefficient & Std. Error & $p$ \\ 
  \hline
\textit{Fixed effects}& Intercept & 22.553 & 2.295   & $8.38 \times 10^{-16}$ \\ 
  & AOI guidance & $-6.237$ & 2.720  & 0.022 \\ 
  & post-guidance control & $-11.351$ & 2.806  & $6.22 \times 10^{-5}$ \\ 
  & GT = wAMD & 8.847 & 1.091  & $5.75 \times 10^{-15}$ \\ 
  & within-block position & $-0.915$ & 0.212  & $1.92 \times 10^{-5}$ \\ 
  & AOI guidance:within-block position & 0.585 & 0.299 & 0.051 \\ 
  &post-guidance control:within-block position & 0.595 & 0.308  & 0.054 \\ 
     \hline
 & Group & Name & Variance & Std. Dev. \\ 
   \cline{2-6}
    \textit{Random effects} &Participant&Intercept&13.01&3.607\\
    & Residual & & 125.25 & 11.91& \\
      \hline
       & &$R^2_{model}$ & $R^2_{fixed}$ & $R^2_{random}$& \\
       \cline{2-6}
      \textit{Model fit}&&0.263&0.188&0.075&\\
      &&&&&\\
\textbf{VLM experiment} & Variable & Coefficient & Std. Error  &$p$  \\ 
  \hline
  
\textit{Fixed effects}& Intercept & 26.478 & 6.333   & $8.44 \times 10^{-4}$ \\ 
  & VLM guidance & $-8.320$ & 6.691  & 0.216 \\ 
  & GT = wet AMD & 13.036 & 3.296   &  $1.34 \times 10^{-4}$ \\ 
  &within-block position & $-0.172$ & 0.521  & 0.742 \\ 
  & VLM guidance:within-block position & 0.636 & 0.735 & 0.389 \\ 
   \hline
    & Group & Name & Variance & Std. Dev. \\ 
   \cline{2-6}
    \textit{Random effects} &Participant&Intercept&60.98&7.81\\
   &  Residual & & 302.80 & 14.40\\
   \hline       & &$R^2_{model}$ & $R^2_{fixed}$ & $R^2_{random}$& \\
       \cline{2-6}
      \textit{Model fit}&&0.246&0.107&0.139&\\
\end{tabular}
    \caption{Mixed effects modeling of time spent per eye in each experimental condition of \textbf{US2}. ``GT" = Ground Truth, ``:" = interaction between.}
    \label{tab:mixed-time}
    \Description{A table reporting mixed-effects linear models of time spent per eye in US2 for the AOI and VLM experiments, listing fixed-effect coefficients, standard errors, and p-values for the intercept, the guidance and post-guidance-control blocks, ground-truth wAMD, within-block position, and interactions; by-participant and residual random effects; and R-squared model-fit values. In the AOI model, AOI guidance reduces time per eye by 6.24 seconds (p=0.022) and post-guidance control by 11.35 seconds (p below 0.001), while wAMD cases take longer; in the VLM model the guidance effect is non-significant (p=0.216).}
\end{table*}

\clearpage
\section{VLM Comment Analysis}

\begin{figure}[t]
    \centering
    \includegraphics[width=0.5\linewidth]{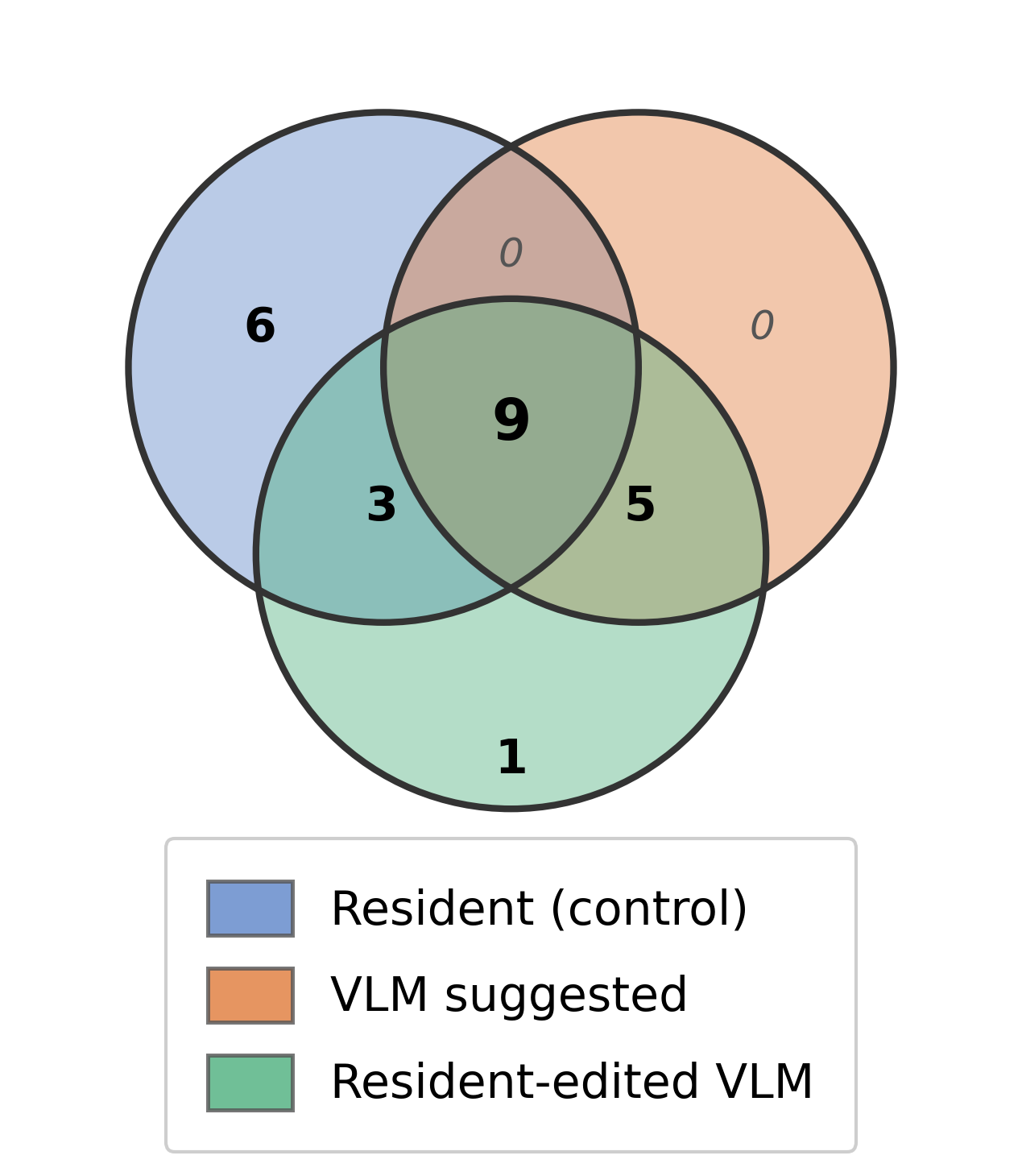}
    \caption{Overlap of unique identified biomarkers across all VLM user study conditions.}
    \label{fig:vlm_venn_supplement}
    \Description{A three-way Venn diagram showing overlap of unique biomarker vocabulary across three sets: Resident (control block, blue circle), VLM suggested (orange circle), and Resident-edited VLM (green circle). Numbers in each region show unique biomarker counts: Resident-only = 6, VLM-only = 0 (grey italic), Resident-edited VLM only = 1, Resident and Resident-edited overlap = 3, VLM and Resident-edited overlap = 5, Resident and VLM only = 0 (grey italic), all three = 9. A legend below identifies each circle by color.}
\end{figure}

\autoref{fig:vlm_venn_supplement} shows vocabulary overlap across three sets: VLM-provided biomarkers, user-provided biomarkers (control block), and user-modified VLM-provided biomarkers.

\section{User Survey Questions}

The following items were queried during the post-experiment surveys in US2.

\textbf{Participant Demographics}

\begin{enumerate}
    \item Participant level of experience
    \item Participant confidence in reading OCT scans [1,4]
\end{enumerate}

\textbf{General UI \& Control Comments}

\begin{enumerate}
    \item What were your impressions of the system's interface and usability?
    \item Any comments about your experience interpreting the images in the first block today?

\end{enumerate}

\textbf{AOI-specific Questions}

\begin{enumerate}

    \item How helpful was the guidance provided by the heatmaps? [1,5]
    \item Did the heatmap guidance assist you in identifying potential cases of wet AMD more efficiently? Please provide specific examples or scenarios if possible.
    \item Do you have any suggestions on how the guidance could be improved?
    \item Which of the following changes to the guidance method do you feel could improve its ability to help you interpret OCT scans? (A graphical change in how the heatmaps are presented, Heatmaps with fewer highlighted areas, Heatmaps with more highlighted areas, An auto-generated text summary of the OCT images,  Ability to change the heatmap in real time)
    \item Any additional comments or feedback you would like to provide about the system and experiment?
\end{enumerate}

\textbf{VLM-specific Questions}

\begin{enumerate}
    \item How would you rate the accuracy of the text summaries? [1,5]
    \item What comments do you have about the accuracy of the text summaries?
    \item How did your workflow change when text summaries were provided?
    \item Could you see yourself using a system that auto-generates text summaries in clinical practice? Why or why not?
    \item How would the following changes to the text prompt method change its ability to help you interpret OCT scans? [1,5]: More succinct descriptions; More verbose descriptions; A confidence score on the description; Highlighted areas of relevance from which the descriptions are drawn; A conversational AI agent to interactively clarify and update the description.
    \item Any additional comments or feedback you would like to provide about the text system and experiment?
    
\end{enumerate}

\textbf{Post-Guidance Questions}

\begin{enumerate}
    \item In the unguided block (after the guided block), what difference did you feel in reading the OCT images?
    \item To what extent did you want the guidance back? [1,5]
    \item How do you feel about your efficiency in reading the OCT images in the unguided condition again?
\end{enumerate}

\majoredit{\section{US3 Post-Study Survey Questions}}

\majoredit{The following items were queried during the post-experiment survey in US3 (combined VLM+AOI guidance), administered after all three blocks.}

\majoredit{\textbf{Likert Items (1--7 scale)}}

\begin{enumerate}
    \item \majoredit{The guidance was accurate enough that I felt comfortable acting on it.}
    \item \majoredit{Compared to earlier blocks, I relied on the on-screen guidance when deciding what to do.}
    \item \majoredit{Compared to the block right before this one, I felt less supported.}
    \item \majoredit{Even without guidance in this block, I used things I learned from the earlier block.}
    \item \majoredit{If I could choose, I would want the guidance from the earlier block available here.}
\end{enumerate}

\majoredit{\textbf{Open-Ended Questions}}

\begin{enumerate}
    \item \majoredit{Describe a moment when the guidance felt helpful.}
    \item \majoredit{Describe a moment when the guidance felt unhelpful.}
    \item \majoredit{After the guidance was removed, what (if anything) did you do differently comparing to before seeing the guidance?}
    \item \majoredit{If you could change one thing about the guidance to better fit your workflow, what would it be?}
\end{enumerate}

\begin{figure*}[tp]
  \centering
  \includegraphics[width=\textwidth]{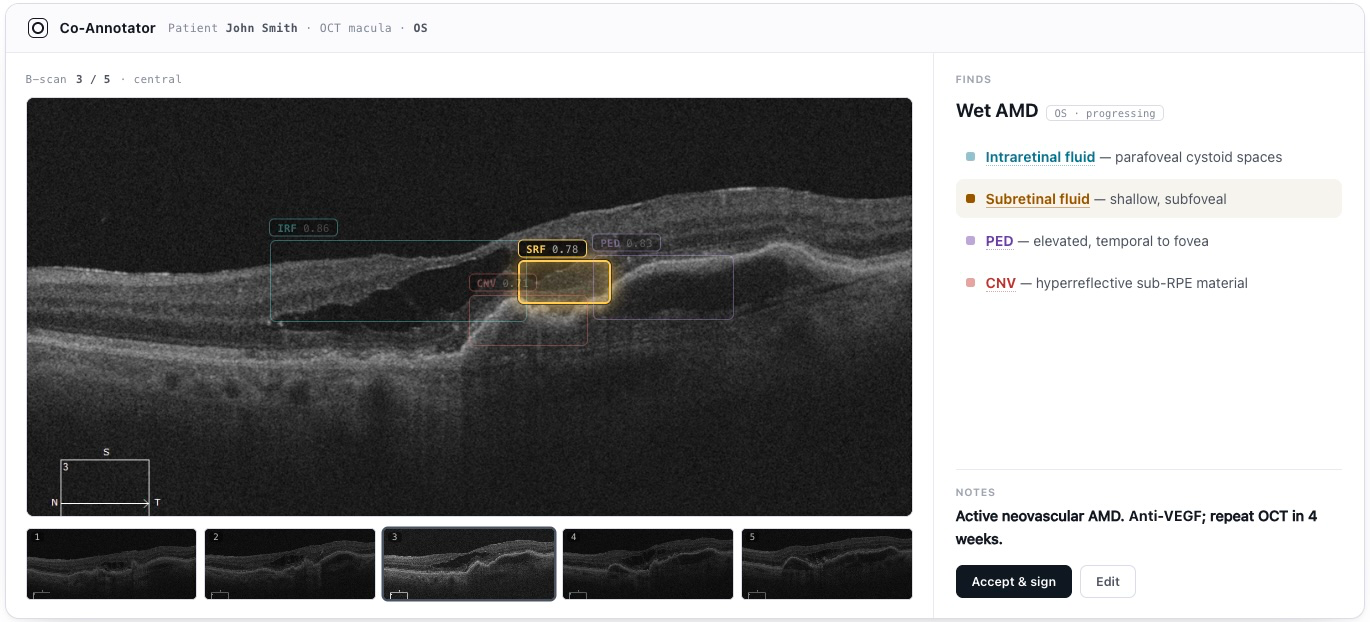}
  \caption{\majoredit{Proposed per-biomarker, evidence-anchored guidance (design mock-up). Hovering or selecting a biomarker token in the findings panel (here, subretinal fluid) highlights only that biomarker's feature region on the active B-scan, replacing the global AOI heatmap; each labeled finding is clickable and color-matched to its region. Residents requested this per-biomarker linking in US3. \textbf{Biomarker locations in this figure are purely illustrative.}}}
  \Description{A mock-up of the Co-Annotator interface showing per-biomarker, evidence-anchored highlighting. Left: an OCT B-scan (slice 3 of 5, central) with individually labeled, color-coded regions for intraretinal fluid (IRF, 0.86), subretinal fluid (SRF, 0.78), pigment epithelial detachment (PED, 0.83), and choroidal neovascularization (CNV), each with a confidence value; the selected subretinal-fluid token highlights only its region in yellow rather than showing a global heatmap. Right: a findings panel listing a Wet AMD diagnosis (progressing) and the four biomarkers as clickable, color-matched tokens with short descriptions, plus an editable clinical note and Accept-and-sign / Edit controls. Biomarker locations are illustrative.}
  \label{fig:biomarker-aoi-mockup}
\end{figure*}

\majoredit{\section{Proposed Biomarker-to-AOI Linking}
Motivated by residents' requests in US3 (\autoref{sec:us3-qual}), \autoref{fig:biomarker-aoi-mockup} shows a mock-up of per-biomarker, evidence-anchored guidance: selecting a biomarker token in the findings panel highlights only that biomarker's feature region on the active B-scan, replacing the global AOI heatmap.}

\end{document}